%% file: arxivversion.tex
\documentclass{article} 
\usepackage{iclr2024_conference,times}

\input{math_commands.tex}

\usepackage{hyperref}
\usepackage{amsmath}
\usepackage{url}
\usepackage{listings, booktabs}
\usepackage{graphicx}

\usepackage{wrapfig}
\usepackage[linesnumbered,ruled,vlined]{algorithm2e}
\usepackage{algorithmic}
\usepackage{multicol, multirow}
\usepackage{xcolor}
\usepackage{tabularx}
\usepackage{booktabs}
\usepackage[labelfont=bf,format=plain,justification=raggedright]{caption}
\usepackage{arydshln}

\SetKwBlock{DoParallel}{DoParallel}{end}
\SetCommentSty{mycommfont}
\usepackage{amsthm}

\theoremstyle{definition}

\usepackage{soul}

\definecolor{codegreen}{rgb}{0,0.6,0}
\definecolor{codegray}{rgb}{0.5,0.5,0.5}
\definecolor{codepurple}{rgb}{0.58,0,0.82}
\definecolor{backcolour}{rgb}{0.95,0.95,0.92}

\lstdefinestyle{mystyle}{
  backgroundcolor=\color{backcolour}, commentstyle=\color{codegreen},
  keywordstyle=\color{magenta},
  numberstyle=\tiny\color{codegray},
  stringstyle=\color{codepurple},
  basicstyle=\ttfamily\footnotesize,
  breakatwhitespace=false,         
  breaklines=true,                 
  captionpos=b,                    
  keepspaces=true,                 
  numbers=left,                    
  numbersep=5pt,                  
  showspaces=false,                
  showstringspaces=false,
  showtabs=false,                  
  tabsize=2
}

\usepackage[strict]{changepage}
\usepackage{xcolor}
\usepackage{framed}
\definecolor{demonstrationshade}{rgb}{0.95,0.95,1}
\definecolor{promptshade}{rgb}{0.95,0.95,1}

\usepackage{authblk}

\title{From Passive Delegates to Strategic Negotiators:\\Reinforcing Social Reasoning in Small Language Models with SocialRL}

\author{{Wenyue Hua}\thanks{Corresponding authors: wenyuehua@microsoft.com, aslic@microsoft.com, samershi@microsoft.com}}
\author{{Zachary Huang}}
\author{{Tyler Payne}}
\author{{Safoora Yousefi}}
\author{{Saleema Amershi}}
\author{{Asli Celikyilmaz}}
\affil{Microsoft Research, AI Frontiers}

\begin{document}

\maketitle

\begin{abstract}
AI agents increasingly act on their users' behalf as representatives of their interests, handling tasks such as scheduling meetings, comparing offers, and haggling over prices. These principal-driven tasks routinely place the agent across from a counterpart, such as another user's agent, a seller, or a recruiter, whose goals may conflict with those of its principal. Yet the dispositions that make an assistant pleasant can make it a poor delegate: a friendly and helpful frontier model may disclose its principal's private information unprompted, and concede the principal's position at the first sign of resistance. We present \textsc{SocialRL}, a general recipe that trains social reasoning directly, and apply it to a 4B model across six interaction domains: Deal-or-No-Deal, CaSiNo, Craigslist, Job Interview, Calendar, and Marketplace. Every domain is trained in-domain under the same recipe, and every trained policy is evaluated on all six domains. We find that (1) in-domain training reaches the frontier: on held-out scenarios the 4B matches or exceeds the GPT-5 family per domain, closing 73--122\% of the baseline-to-frontier gap on the negotiation games, with the change visible at the trace level where 78\% of buyer openings anchor below target versus 3\% untrained; (2) cross-domain transfer follows game structure: structurally paired games lift each other, a broad multi-issue donor lifts nearly all domains, and structurally isolated games transfer nothing; (3) guided by this transfer structure, we propose two strategies to consolidate the per-domain specialists into a single unified 4B that achieves 0.627 average utility across all six environments, matching or exceeding GPT-4.1 (0.625), GPT-5.1 (0.619), GPT-5.2 (0.613): cascade RL and multi-teacher on-policy distillation (OPD); (4) an explicit theory-of-mind (ToM) scaffold helps only through training: distilling the ToM trace, rather than actions alone, lifts utility on every environment and generalizes better across them, and of the two ToM skills, only next-action prediction predicts negotiation outcomes. 
\end{abstract}

\section{Introduction}
AI agents are no longer only tools for completing tasks. A growing share of what they are asked to do carries the user's stake: the agent acts on the user's behalf, and the outcome bears directly on the user's interests \citep{sun2025game, south2025authenticated}. Deployed systems can manage customer-service requests, coordinate meetings, purchase goods and book reservations, and help users compare homes or schedule property tours \citep{salesforce_agentforce, google_schedule, openai_operator, zillow_ai}. These applications position an AI agent as a \emph{delegate}: a system entrusted with a principal's preferences and authorized to take consequential actions in open-ended environments.

Delegation becomes especially challenging when an agent interacts with another party whose objectives differ from those of its principal. A customer-service agent~\citep{chaturvedi2023opportunities} may face a user seeking compensation, a purchasing agent may negotiate with a seller seeking a higher price~\citep{haurum2024real}, and a scheduling agent~\citep{zou2026calbench} may need to reconcile participants with competing constraints. In such settings, merely completing the interaction is insufficient. A capable delegate must protect private information~\citep{juneja2025magpie}, infer the counterpart's incentives~\citep{matta2026artificial}, determine when to concede or push back, and pursue an agreement that advances its principal's interests~\citep{sun2025llm}. We refer to this collection of capabilities as \emph{social reasoning}: reasoning about another actor's latent preferences, incentives, and likely behavior in order to choose strategically effective actions.

\begin{figure}
    \centering
    \includegraphics[width=0.8\linewidth]{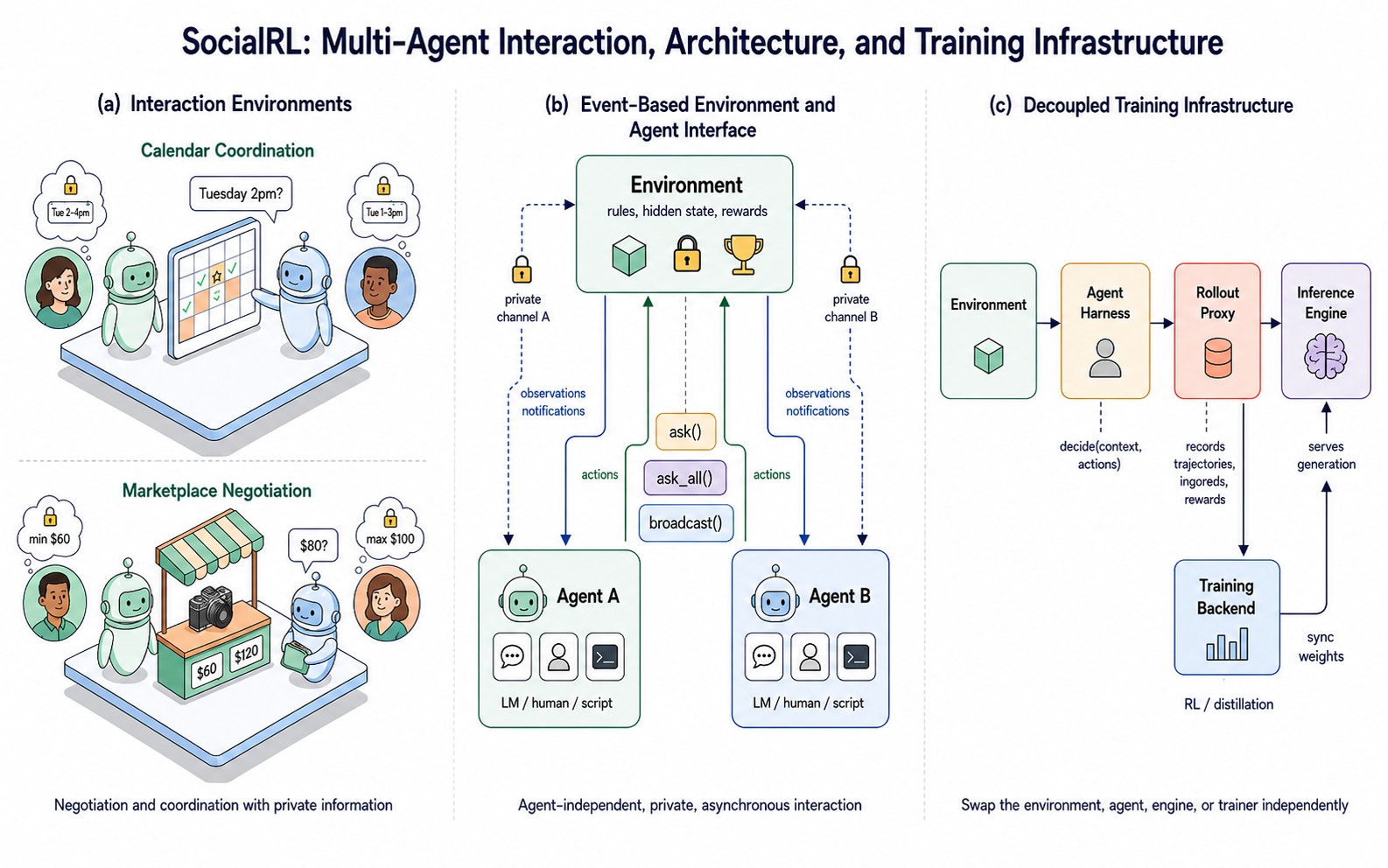}
    \caption{\textbf{SocialRL overview: interaction environments, event-based agent interface, and decoupled training infrastructure.} \textbf{(a)} Example social-reasoning environments for delegated coordination and negotiation. \textbf{(b)} An event-based, agent-independent interface in which a stateful environment communicates with agents through private channels of observations, notifications, and actions. \textbf{(c)} A decoupled training stack where the rollout proxy connects the environment and agent harness to inference and training, enabling interchangeable components and supporting both reinforcement learning and distillation.}
    \label{fig:main}
\end{figure}

Negotiation has long served as a testbed for language-based strategic interaction, including multi-item allocation, resource division, and price bargaining \citep{lewis2017deal, he2018decoupling, chawla2021casino}. Recent work has extended these settings to evaluate the agency and strategic behavior of general-purpose language models \citep{bianchi2024well, zhang2026terms, hua2024game}. Although frontier models can conduct coherent negotiations and frequently reach agreements, strong aggregate outcomes do not guarantee faithful representation of the principal. We find that frontier models routinely abandon contested positions after limited resistance. These behaviors reflect dispositions that are useful for general assistance, such as transparency, agreeableness, and an eagerness to reach consensus, but they leave an agent predictable and exploitable when acting as a delegate.

These failures reflect a mismatch between the objectives used to train general-purpose assistants and those required for strategic delegation. Instruction tuning and reinforcement learning from human feedback are typically designed to produce broadly helpful~\citep{dahlgren2025helpful}, honest, harmless, and instruction-following behavior in cooperative dialogue~\citep{kirk2024understanding,zhou2023instruction,dong2024rlhf}. In interactions with conflicting objectives and private information, the same behavioral priors can manifest as premature disclosure, excessive accommodation, and a preference for agreement even when the resulting outcome is unfavorable to the principal. Delegated agency therefore requires principal-conditioned strategic behavior: selective information disclosure, calibrated reservation boundaries, and the willingness to reject or prolong an interaction when agreement would sacrifice the principal's utility. This objective mismatch motivates post-training that targets social reasoning directly.

We present \textsc{SocialRL}, a complete infrastructure + recipe for training and studying social reasoning in language-model agents, as shown in Figure\ref{fig:main}. We focus on a 4B model to test whether strategic delegation can be induced through targeted post-training rather than depending on frontier-scale capacity. Using \textsc{SocialRL}, the resulting 4B policies approach and, in some settings, exceed the performance of much larger GPT models while learning strategies that generalize across interaction structures. We train and evaluate across six interaction domains: Deal-or-No-Deal~\citep{lewis2017deal}, CaSiNo~\citep{chawla2021casino}, Job Interview~\citep{yamaguchi2021dialogue}, Craigslist~\citep{he2018decoupling}, and Calendar and Marketplace from SocialReasoning-Bench~\citep{srbench2026}. Together, these domains span single- and multi-issue bargaining, price negotiation~\citep{zhu2025automated}, and slot coordination. To support training in these multi-turn environments, we build a decoupled system~\citep{zeng2025glm, wang2026openclaw, luo2025agent} that separates the environment, agent harness, inference engine, and trainer, allowing a common interaction infrastructure to support reinforcement learning, distillation, and heterogeneous or black-box counterparts. We first train a specialist within each domain and evaluate the full cross-environment transfer matrix, then consolidate the specialists into a single policy across all six environments using cascade RL~\citep{du2024cascading} as well as multi-teacher mode-seeking OPD.

On held-out scenarios, our domain-trained 4B policies achieve performance comparable to much larger GPT models across the six environments. Cross-environment transfer is substantial and follows interaction structure, with the strongest transfer occurring between structurally similar domains. We consolidate these specialized policies through cascade reinforcement learning and multi-teacher OPD (MOPD), producing a unified 4B model that achieves 0.627 average utility across all six environments, comparable to GPT-4.1 at 0.625, GPT-5.1 at 0.619, GPT-5.2 at 0.613. Finally, we introduce explicit theory-of-mind supervision through \textsc{Infer} $\to$ \textsc{Act} $\to$ \textsc{Anticipate}. Distilling these reasoning traces improves performance and cross-environment generalization, with next-action prediction emerging as the theory-of-mind skill most predictive of negotiation outcomes.

\paragraph{Contributions.}
\begin{itemize}

\item \textbf{Strategic Delegation Formulation \& Six-Environment Suite.}
We formulate strategic delegation as a social-reasoning post-training problem and introduce a heterogeneous suite spanning allocation, multi-issue bargaining, price negotiation, and preference-based coordination. We train domain specialists in each environment and evaluate every policy across the full cross-environment transfer matrix.

\item \textbf{Agent-Agnostic Environments \& Decoupled Training Infrastructure.}
We develop \textsc{SocialRL}, combining an event-based, agent-independent environment interface with an OpenAI-compatible rollout proxy that disentangles environments, agent harnesses, inference engines, and trainers. The same stack supports reinforcement learning and distillation with local, remote, heterogeneous, or black-box counterparts.

\item \textbf{Frontier-Range 4B Policies \& Transfer-Aware Unification.}
We show that domain-trained 4B policies achieve aggregate performance in the range of GPT-4.1, GPT-5.1, and GPT-5.2, and reveal that cross-environment transfer is strongly directional and structure-dependent. We exploit this transfer structure to consolidate the specialists: transfer-aware cascade RL reaches $0.627$ Avg-6, while multi-teacher on-policy distillation recovers $92.6\%$ of the specialists' average advantage in only $60$ additional optimization steps.

\item \textbf{Explicit Theory-of-Mind Supervision.}
We introduce \textsc{Infer} $\rightarrow$ \textsc{Act} $\rightarrow$ \textsc{Anticipate} supervision and show that distilling the complete reasoning trace outperforms action-only supervision on every evaluated negotiation environment and improves cross-environment generalization. We further identify next-action prediction, rather than preference inference alone, as the theory-of-mind component most predictive of negotiation outcomes.

\end{itemize}

\section{Related Work}
\label{sec:related}

\paragraph{Language-model agents.}
A large body of work on language-model agents has focused on enabling models to reason, plan, and act through external tools and digital interfaces. ReAct~\citep{yao2022react} introduced the interleaving of language-based reasoning with environment actions, while Toolformer~\citep{schick2023toolformer} demonstrated that language models can learn when and how to invoke external APIs. Subsequent benchmarks have evaluated increasingly realistic forms of web navigation, desktop control, and tool-mediated interaction, including WebArena, OSWorld, and $\tau$-bench \citep{zhou2023webarena, xie2024osworld, yao2024taubench}. This literature has substantially advanced planning~\citep{wang2026agent}, tool selection~\citep{yang2026evotool}, interface grounding~\citep{awadallah2025fara}, and policy compliance~\citep{elkoussy2026agentltl}, with performance generally measured by whether the agent reaches a target environment state. Strategic interaction introduces an additional dimension: the environment contains another adaptive decision-maker~\citep{anantaprayoon2026learning}, so the value of an action depends on its immediate effect, the information it reveals, and the future responses it induces.

\paragraph{LLMs as delegated agents.}
LLMs are increasingly studied as delegates that act on behalf of users or organizations in economically and socially consequential settings. $\tau$-bench\citep{yao2024taubench} models agents handling retail and airline customer-service requests under domain-specific policies, while ScheduleMe~\citep{anantaprayoon2026learning} applies multi-agent coordination to personal calendar management. CalBench further studies calendar assistants that coordinate under private information, exposing trade-offs between scheduling efficiency, fairness, communication, and privacy \citep{zou2026calbench}. In commerce, ACES evaluates agents that inspect marketplaces and make product choices on behalf of consumers \citep{allouah2026your}; in negotiation, recent work has compared advisory, coaching, and autonomous delegation interfaces in multi-party bargaining. These settings require the agent to preserve and act on a principal's preferences while interacting with users, platforms, or other agents whose objectives may differ. Negotiation is a particularly direct instance of delegated agency because success depends jointly on reaching an agreement, managing private information, and securing value for the represented principal.

\paragraph{Training agents for delegated interaction.}
A growing line of work trains language agents through strategic interaction. Within negotiation, prior work has explored self-play and language feedback, iterative self-play with behavior cloning, reinforcement learning with verifiable economic rewards, and pipelines combining supervised training on synthetic negotiations with reinforcement learning~\citep{fu2023improving,liao2024efficacy,liu2026instructing,bergemann2026training}. These methods can induce substantially stronger bargaining strategies, but typically specialize to a particular interaction structure, such as bilateral price bargaining or resource division. Complementary work has broadened the scope of interactive post-training: \textsc{Sotopia-$\pi$} and \textsc{Sotopia-RL} train general social behavior from open-ended interactions~\citep{wang2024sotopia,yu2025sotopia}, while recent self-play methods train transferable multi-agent reasoning across cooperative and competitive games~\citep{yuan2025mars,jiang2026one,lyu2026gift}. Evaluation frameworks have likewise highlighted the diversity of negotiation itself, spanning resource allocation, exchange, price bargaining, and realistic multi-issue scenarios~\citep{bianchi2024well,zhu2026piearena}. 

\textsc{SocialRL} focuses on the intersection of these directions: post-training a single small model across heterogeneous delegated interactions that demand different strategic capabilities but share a common need to reason about counterpart incentives and act on behalf of a principal. We study transfer across these interaction structures and show that the resulting 4B policies reach the performance range of much larger GPT models.

\section{Environment and Infrastructure for SocialRL}

\begin{figure}[!ht]
    \centering
    \includegraphics[width=0.7\linewidth]{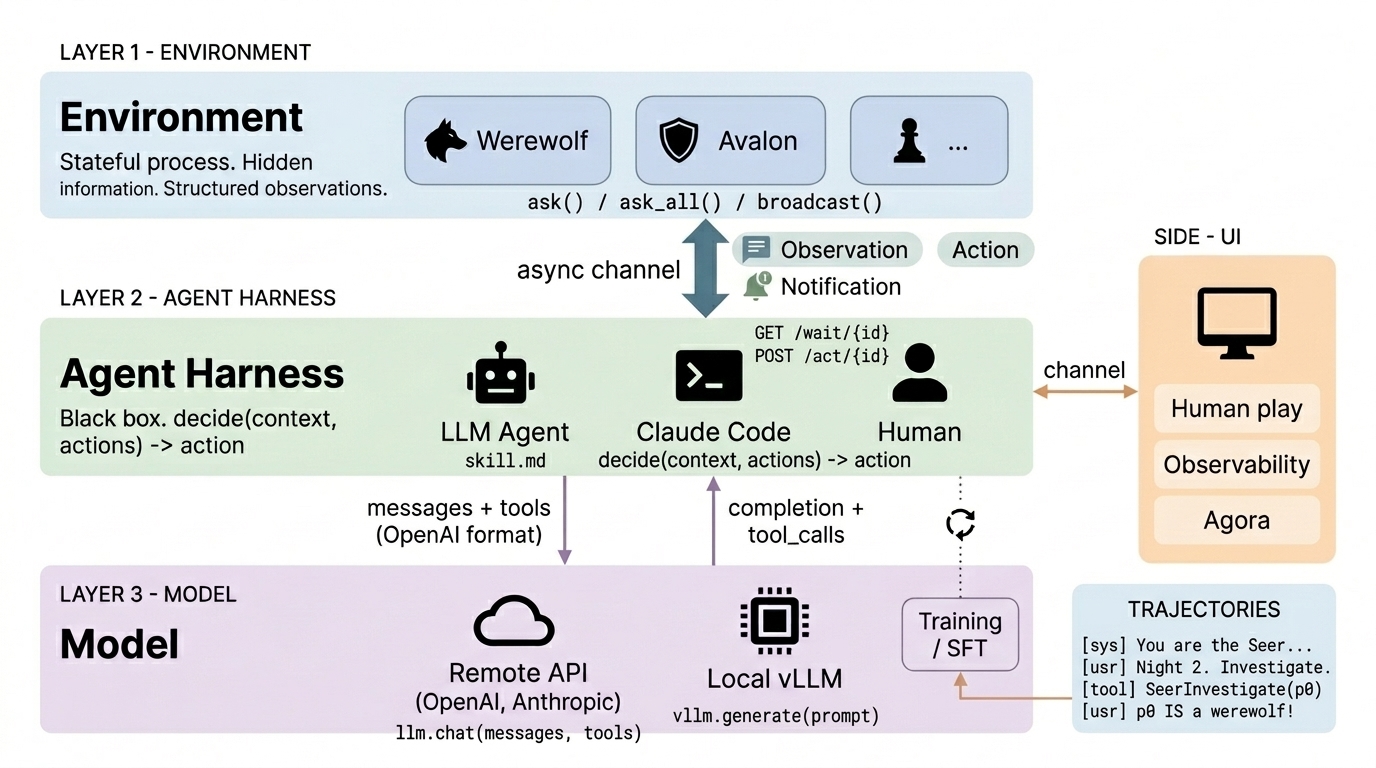}
    \caption{\textbf{Environment and agent architecture.}
    A stateful environment communicates with black-box agents through per-agent asynchronous channels carrying structured observations, notifications, and actions. The same interface supports LLM agents, external coding agents, humans, and scripted policies. Model inference is isolated behind an OpenAI-compatible interface, while the event stream supports agent execution, human play, observability, replay, and trajectory collection.}
    \label{fig:environment-architecture}
\end{figure}

\begin{figure}[!ht]
    \centering
    \includegraphics[width=0.9\linewidth]{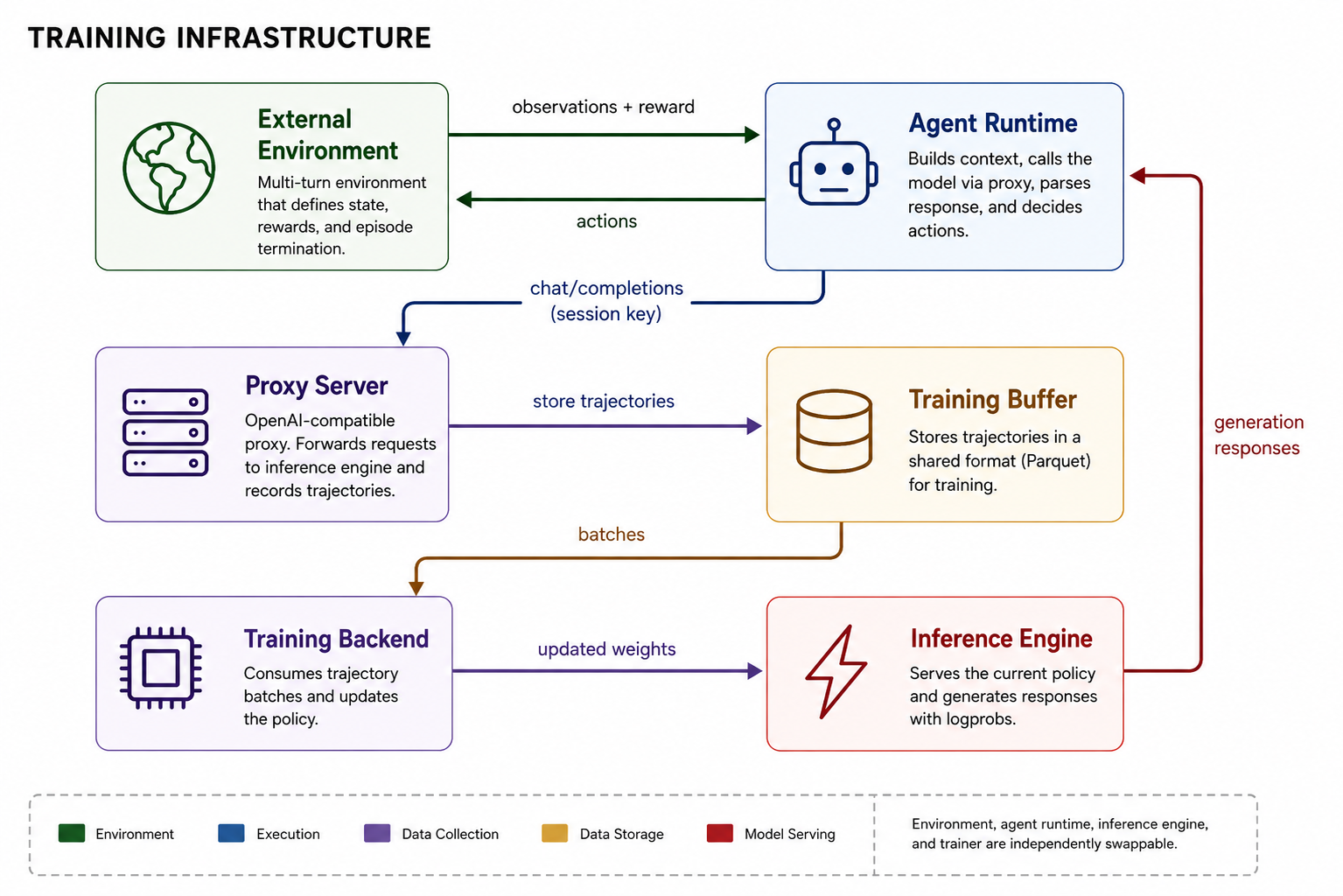}
    \caption{\textbf{Decoupled training architecture.}
    The agent runtime executes independently of the trainer and sends standard model requests through a rollout proxy. The proxy forwards requests to the inference engine while recording the authentic model inputs and outputs associated with each episode. Completed trajectories are written to a shared Parquet buffer consumed by interchangeable training backends, and updated policy weights are synchronized back to the inference engine.}
    \label{fig:training-architecture}
\end{figure}

Training agents through negotiation requires infrastructure that differs from conventional single-turn language-model post-training. An environment may be stateful, partially observed, and populated by multiple agents acting on different schedules. Each episode contains several model calls, while the reward is often available only after the complete interaction. In addition, experiments may combine local trainable policies, remote frontier-model opponents, scripted agents, and human participants. We therefore design the system around two forms of decoupling: an event-based interface separates environments from agents, and an OpenAI-compatible proxy separates agent execution from model training. Figure~\ref{fig:environment-architecture} shows the environment stack, while Figure~\ref{fig:training-architecture} shows the training data flow.

\subsection{Environment and Agent Interface}
\label{sec:environment-design}

We design a generic interface for multi-agent interaction that supports sequential and simultaneous decisions, asynchronous communication, and arbitrary numbers of participants. The same abstraction can represent two-party negotiation environments such as Deal-or-No-Deal and Craigslist, as well as more complex multi-party environments such as Werewolf and Avalon. The design follows four principles: environments are independent of agent implementation; interaction is represented as an event stream with explicit action semantics; agents consume this stream asynchronously; and failures are handled without terminating the episode.

\paragraph{Agent-independent environments.}
The environment is a stateful process that owns the game rules, hidden state, legal actions, and reward computation, while making no assumptions about how participants are implemented. Each participant interacts with the environment through a private asynchronous channel and is treated as a black box implementing a single operation, \texttt{decide(context, actions) $\rightarrow$ action}. An agent may therefore be backed by a language model, an external coding agent, a scripted policy, or a human interface. We further separate the model layer from the agent harness: model inference exposes only standard chat messages and tool definitions through an OpenAI-compatible API. This separation allows the environment, agent implementation, model backend, and execution transport to vary independently.

\paragraph{Event-based interaction and action semantics.}
Communication between the environment and an agent is represented as an ordered stream of events rather than repeated snapshots of environment state. We distinguish two event types: an \emph{observation} records something that has occurred, while a \emph{notification} requests a decision from the receiving agent and specifies its currently available actions. The environment drives interaction through three primitives: \texttt{ask()} for sequential decisions, \texttt{ask\_all()} for simultaneous decisions, and \texttt{broadcast()} for events requiring no response. The control flow of the environment coroutine therefore expresses the game logic directly, without requiring a separate state-machine or dispatcher abstraction.

An event stream preserves information that is important in partially observable multi-agent interaction: what happened, when it happened, and which participants observed it. It also naturally separates an agent's \emph{action} from its \emph{effect}. Submitting an action records a decision and returns only an acknowledgment or validation error; its consequences are emitted subsequently as observations. This distinction is important because actions need not map one-to-one to outcomes. An outcome may depend on several participants' actions, occur only after additional environment logic, or be visible to different participants in different ways. For example, a vote is an individual player's action, whereas an elimination is a collective outcome produced after all votes have been received.

Partial observability is enforced at this same boundary. Each emitted event is filtered for its intended audience, so an agent's channel contains exactly the information available to that participant. The resulting event history can be shared by agent execution, player-facing interfaces, replay tools, and post-episode analysis, reducing the need for separate state projections or logs that may diverge from what the agent actually observed.

\paragraph{Asynchronous event consumption.}
For local execution, agents consume their event streams directly through asynchronous iterators. This interface matches the execution pattern of multi-agent interaction: events arrive incrementally, agents may spend different amounts of time computing their decisions, and progress may depend on responses from other participants. Asynchronous consumption allows agents and the environment to execute concurrently while preserving the ordering of events, without requiring polling or coupling agent execution to the environment's control flow.
Remote agents access the same logical stream through a cursor-based HTTP interface. Events are stored in an append-only history and indexed by sequence number, allowing a disconnected client to resume from its last cursor without losing observations. Local and remote execution therefore share the same interaction semantics and differ only in transport.

\paragraph{Failure and concurrency handling.}
The interface is designed so that transient agent and transport failures do not invalidate an episode. Every submitted action identifies the notification to which it responds. If the environment has advanced while an agent was generating its response, the action is rejected as stale and the agent consumes the newer events before deciding again. Together with the append-only event history, this provides resumable event delivery while ensuring that at most one action is accepted for each decision point.

Invalid actions, timeouts, and unavailable agents are handled explicitly by the environment. Invalid actions generate informative events and may be retried, while a timeout applies an inert default action and records the failure in the event stream. A malformed response, failed model request, or dropped connection therefore becomes part of the trajectory rather than terminating the episode. This property is particularly important for large-scale rollout collection, where a single interaction may involve several independently executing models and individual requests may fail or exceed their latency budgets.

\subsection{Decoupled Training Infrastructure}
\label{sec:training-infrastructure}

Our second design goal is to disentangle rollout generation from training. Agentic rollouts may involve arbitrary environment logic, agent harnesses, memory systems, tool use, and remote counterparts, while the trainer should only require the model inputs, model outputs, and eventual rewards needed to optimize the policy. We therefore connect rollout and training through a single abstraction: an OpenAI-compatible \emph{rollout proxy} placed between the agent runtime and the inference endpoint.

\paragraph{Rollout-training separation.}
The environment and agent harness execute exactly as they would at inference time. The environment determines observations, actions, and rewards; the agent decides how to construct prompts, manage memory, invoke tools, and interpret model outputs. Neither contains training-specific logic. From the trainer's perspective, these implementation details are invisible: it consumes trajectories captured at the model API boundary and therefore need not know how the environment or agent harness is implemented.

This separation is particularly useful for agentic interaction, where a single episode may contain many model calls, the terminal reward may become available only after the interaction ends, and counterparts may be remote or black-box systems outside the training process. It also allows the rollout stack and training stack to evolve independently.

\paragraph{The rollout proxy as the training boundary.}
The proxy exposes the same OpenAI-compatible interface as the underlying inference endpoint, so an agent can use it by changing only the client's base URL. For every model call, the proxy records the exact request received from the agent and the exact response returned by the model, while forwarding the response to the agent unchanged. It therefore captures the authentic model inputs and outputs produced during execution, without requiring the harness to construct a separate training representation.

Each episode is associated with a unique session key carried with every model request. The session groups otherwise stateless HTTP calls into a trajectory. When an episode begins, the runtime opens a session; once the interaction terminates, it attaches the final reward and closes the session. Many rollout workers can consequently execute concurrently against a shared continuously batched inference server without exposing their internal control flow to the trainer.

\paragraph{Trajectory reconstruction and memory compaction.}
Importantly, the proxy does not assume that the agent maintains its context in any particular way. Agent harnesses commonly truncate histories, summarize earlier interactions, or otherwise compact memory during long rollouts. In such cases, a later model request may no longer extend the prefix of an earlier request.
We handle this directly at the model-call boundary. Consecutive calls whose contexts share the expected prefix are treated as part of the same training segment. When this prefix relation breaks, the proxy starts a new segment. Thus, context truncation, summarization, and memory compaction require no special integration with the trainer: they simply induce multiple training segments within the same episode. Each segment retains the exact context under which its response was generated, avoiding any need to reconstruct an artificial conversation history that the policy never actually observed.

\paragraph{A common contract for RL and distillation.}
The proxy supports both reinforcement learning and distillation while preserving the same rollout interface. For reinforcement learning, it records tokenized prompts and responses together with token-level log probabilities from the generating policy. Capturing these probabilities at generation time preserves the behavior policy required by PPO even if the policy is updated before optimization. For distillation, the proxy stores the raw messages, tool definitions, responses, and tool calls, allowing trajectories from closed-source or heterogeneous teachers to be tokenized later using the student's tokenizer.

Completed sessions are exported to a shared Parquet buffer that forms the data contract between rollout and training. RL records include response tokens, rollout log probabilities, rewards, task identifiers, and policy versions; distillation records preserve the raw interaction trace. Training losses are applied only to model response tokens. Because trainers consume this common representation, the rollout system can be paired with a lightweight local trainer, distributed PPO through veRL, alternative algorithms through phitrain, or external training backends through a small adapter.

\paragraph{Asynchronous rollout and optimization.}
Rollout collection, optimization, and inference proceed asynchronously. Episode workers continuously generate trajectories and append completed sessions to the buffer; the trainer consumes them in batches, updates the policy, and synchronizes new weights to the inference server. The rollout and training systems can therefore scale independently.
Asynchrony can introduce policy staleness when rollout generation runs ahead of optimization. We bound this effect with a capacity-based gate that limits the number of trajectories dispatched ahead of the trainer. Each trajectory is tagged with the policy version that generated it, while remaining off-policy differences are handled by the importance-sampling correction used during PPO.
Together with the environment interface in Section~\ref{sec:environment-design}, this design separates the major axes of an agent-training system: environments define interactions, harnesses define agent behavior, the proxy records authentic model calls, and trainers optimize over the resulting trajectories. Changing any one of these components does not require rewriting the others.

\section{Social Reasoning Environments and Reward Design}
\label{sec:datasets}

\textsc{SocialRL} trains agents across six heterogeneous negotiation and coordination environments: Deal-or-No-Deal, CaSiNo, Craigslist Bargains, Job Interview, Calendar, and Marketplace. Our training recipe proceeds in two stages. We first train a domain-specialized policy in each environment, then consolidate these specialists into a single policy that performs well across all six domains. We study two consolidation strategies in Section~\ref{sec:training}: cascade reinforcement learning, whose training order is chosen according to measured cross-domain transfer, and MOPD, which provides a substantially more efficient route to unification.

The six environments are deliberately heterogeneous. They span multi-issue allocation, multi-issue contract negotiation, single-issue price bargaining, and preference-based coordination, with different action spaces, utility structures, information asymmetries, and interaction dynamics. Their common structure is social reasoning: in every domain, an agent must reason about a counterpart with private objectives, protect information about its own principal, and decide when to propose, concede, reject, or close. This diversity lets us study both specialization within individual interaction structures and transfer of social-reasoning strategies across them. Table~\ref{tab:envs} summarizes the six environments.

\begin{table}[t]
\centering
\small
\begin{tabular}{@{}lllll@{}}
\toprule
\textbf{Domain} & \textbf{Structure} & \textbf{Issues} & \textbf{Private information} & \textbf{Roles} \\
\midrule
Deal-or-No-Deal & multi-issue allocation & 3 item types & per-item values & symmetric \\
CaSiNo & multi-issue allocation & 3 resources & priority order & symmetric \\
Craigslist & single-issue price & price & target / listing price & asymmetric \\
Job Interview & multi-issue contract & 5 issues & utility weights & asymmetric \\
Calendar & slot coordination & time slots & slot preferences & asymmetric \\
Marketplace & single-issue price & price & reservation price & asymmetric \\
\bottomrule
\end{tabular}
\caption{\textbf{The six social-reasoning environments.}
The suite spans heterogeneous negotiation and coordination structures while sharing a common requirement to reason strategically about a counterpart with private objectives.}
\label{tab:envs}
\end{table}

\subsection{Interaction Environments}
\label{sec:datasets:envs}

\paragraph{Deal-or-No-Deal (DnD).}
Two agents divide a shared pool of items (books, hats, and balls), each holding private per-item values~\citep{lewis2017deal}. A scenario specifies an item multiset and each player's valuation, with every player's total valuation normalized to $10$ points. Agents alternate free-form messages and structured proposals until a split is accepted, one side walks away, or the round limit is reached. Because valuations are private, effective play requires identifying which items the counterpart values and trading low-value items for high-value ones.

\paragraph{CaSiNo.}
Two campers divide a fixed stock of food, water, and firewood packages, with three units of each resource~\citep{chawla2021casino}. Each player has a private priority ordering over the resources, grounded in a persona and backstory. Priorities map to per-unit utilities (High $=5$, Medium $=4$, Low $=3$), giving each player a maximum score of $36$. CaSiNo shares DnD's multi-issue allocation structure while requiring agents to negotiate through naturalistic needs and justifications.

\paragraph{Craigslist Bargains.}
A buyer and seller negotiate the price of a single listed item~\citep{he2018decoupling}. Scenarios are derived from Craigslist postings across multiple product categories, with private price objectives defining the bargaining range. The environment is one-dimensional and role-asymmetric: the buyer seeks a lower price while the seller seeks a higher one. Effective negotiation therefore depends on anchoring, concession timing, and deciding when the available surplus justifies agreement.

\paragraph{Job Interview.}
A worker and recruiter negotiate a five-issue employment package covering salary, weekly holiday, position, workplace, and company~\citep{yamaguchi2021dialogue}. Each side has private utilities and issue weights, and the parties generally value different dimensions of the contract. Efficient agreements therefore require identifying the counterpart's priorities and exchanging concessions on low-value issues for gains on high-value ones. With roughly $10^4$ possible agreements per scenario, Job Interview provides the largest structured outcome space in the suite.

\paragraph{Calendar.}
Calendar is adapted from SocialReasoning-Bench~\citep{srbench2026}. An agent acts on behalf of a principal to negotiate a meeting time with a requestor whose preferences differ from the principal's. Each side has private preferences over available time slots, and the initial request conflicts with the principal's interests. The agent must gather enough information about the requestor's flexibility while steering the interaction toward a slot favorable to its principal.

\paragraph{Marketplace.}
Marketplace is also adapted from SocialReasoning-Bench~\citep{srbench2026}. A buyer agent negotiates a purchase with a seller while holding a private reservation price. The seller begins from a price unfavorable to the buyer, while a zone of possible agreement is guaranteed to exist. As in Craigslist, success requires strategic price negotiation, but the environment differs in its interface, counterpart behavior, and utility construction, making the pair useful for studying transfer across structurally related domains.

\subsection{Outcome Evaluation and Reward Design}
\label{sec:datasets:reward}

All six environments use terminal, outcome-only rewards: each episode receives a single scalar after agreement, walk-away, or timeout, with no intermediate reward shaping. Our reward design follows two principles. First, every environment maps outcomes to a common $[0,1]$ range, providing consistent endpoints for cross-environment reporting, aggregation, and checkpoint selection during multi-domain training. Second, the reward should reflect the quality of an agreement relative to the opportunities available in the particular scenario and role. The appropriate normalization therefore depends on the interaction structure: DnD, CaSiNo, and Job Interview use scenario- and role-specific reference agreements; Craigslist and Marketplace normalize by the available price-negotiation corridor; and Calendar normalizes by the value range over mutually feasible time slots.

For a fixed scenario and reference point $m$, the transformation is strictly increasing in $z$ and therefore preserves the ranking of outcomes within that scenario. Its purpose is to calibrate how outcomes are valued across scenarios. For example, a normalized utility of $0.6$ may exceed the fair-and-efficient reference in a highly conflicting scenario while falling well below it in a scenario with largely compatible preferences. Centering the reward curve at $m$ assigns the greatest resolution to the transition between weak and strong agreements under the scenario-specific benchmark.

\paragraph{Difficulty-aware rewards for multi-issue negotiation.}
DnD, CaSiNo, and Job Interview evaluate an outcome relative to a scenario-specific reference agreement. This adjustment accounts for variation in the utility that a player can reasonably attain under different combinations of private preferences.

\emph{Allocation games.}
In DnD and CaSiNo, a scenario specifies item counts $\{c_k\}$ and private per-item values $v^a$ and $v^b$. For an allocation $x$, where player $a$ receives $x_k$ units of item $k$, the two players obtain
\begin{equation}
s_a(x)
=
\sum_k x_k v^a_k,
\qquad
s_b(x)
=
\sum_k (c_k-x_k)v^b_k .
\label{eq:allocation-utility}
\end{equation}
The maximum raw score is $S=10$ in DnD and $S=36$ in CaSiNo. Because the allocation spaces are small, we enumerate all feasible outcomes and compute their Pareto frontiers exactly.

We use allocations that are both Pareto-optimal and envy-free to define the reference point. An allocation is envy-free when each player weakly prefers its own bundle to the other player's bundle under its private valuation. For player $i$, we define the \emph{envy-free Pareto maximum}
\begin{equation}
m_i
=
\frac{1}{S}
\max
\left\{
s_i(x):
x \text{ is Pareto-optimal and envy-free}
\right\}.
\label{eq:efmax}
\end{equation}
This reference is the greatest normalized value available to player $i$ among outcomes that remain both efficient and fair. In scenarios for which no allocation satisfies both conditions, we use the Pareto-optimal allocation that minimizes the larger of the two players' envy violations.

\emph{Job Interview.}
Each Job Interview outcome is a deal $d$ assigning one option to each issue. The worker and recruiter receive normalized utilities $u_{\mathrm{w}}(d),u_{\mathrm{r}}(d)\in[0,1]$ derived from their private issue weights and option utilities. Because the complete outcome space is enumerable, we define the reference agreement through the egalitarian objective
\begin{equation}
d^*
=
\arg\max_d
\min\left(
u_{\mathrm{w}}(d),
u_{\mathrm{r}}(d)
\right),
\label{eq:egal}
\end{equation}
breaking ties in favor of greater total utility. The role-specific reference points are
\begin{equation}
m_{\mathrm{w}}=u_{\mathrm{w}}(d^*),
\qquad
m_{\mathrm{r}}=u_{\mathrm{r}}(d^*).
\end{equation}

\emph{Common difficulty-aware transformation.}
Let $z\in[0,1]$ denote the agent's normalized outcome utility and $m\in[0,1]$ its scenario- and role-specific reference point. DnD, CaSiNo, and Job Interview use
\begin{equation}
r
=
\frac{
\sigma\!\left(\frac{z-m}{T}\right)
-
\sigma\!\left(\frac{-m}{T}\right)
}{
\sigma\!\left(\frac{1-m}{T}\right)
-
\sigma\!\left(\frac{-m}{T}\right)
},
\qquad
\sigma(x)=\frac{1}{1+e^{-x}},
\qquad
T=0.2 .
\label{eq:sigmoid}
\end{equation}
The transformation preserves $r=0$ at $z=0$ and $r=1$ at $z=1$, while placing the steepest part of the curve at $z=m$. Outcomes near the reference agreement therefore receive the greatest reward discrimination. The same raw utility can represent strong play in a highly conflicting scenario and a weak agreement in a scenario with largely compatible preferences.

This normalization retains a common absolute reward range while accounting for what was reasonably attainable in each scenario and role. It therefore gives PPO a more comparable signal of agreement quality across scenarios with different utility frontiers.

\paragraph{Price-corridor rewards for single-issue negotiation.}
Craigslist and Marketplace both measure how much of a one-dimensional bargaining surplus each party captures. Let $p_{\min}$ denote the buyer-favorable endpoint of the bargaining corridor, $p_{\max}$ the seller-favorable endpoint, and $p_{\mathrm{deal}}$ the agreed price, with $p_{\min}<p_{\max}$. We define
\begin{equation}
r_{\mathrm{buyer}}
=
\operatorname{clip}
\left(
\frac{p_{\max}-p_{\mathrm{deal}}}
     {p_{\max}-p_{\min}},
0,1
\right),
\qquad
r_{\mathrm{seller}}
=
\operatorname{clip}
\left(
\frac{p_{\mathrm{deal}}-p_{\min}}
     {p_{\max}-p_{\min}},
0,1
\right).
\label{eq:price-corridor}
\end{equation}
For any agreement within the corridor, the two rewards sum to $1$. A deal at $p_{\min}$ assigns the full available surplus to the buyer, while a deal at $p_{\max}$ assigns it to the seller.

In Craigslist, the corridor is defined by the buyer's target price $p_t$ and the listing price $p_{\ell}$:
\begin{equation}
p_{\min}=p_t,
\qquad
p_{\max}=p_{\ell}.
\end{equation}
The resulting reward measures how much of the listing-to-target bargaining range each role captures.

In Marketplace, the corridor is the zone of possible agreement defined by the seller's and buyer's private reservation prices~\citep{srbench2026}. Let $p_s$ be the seller's reservation price and $p_b$ the buyer's reservation price, with $p_s<p_b$. We set
\begin{equation}
p_{\min}=p_s,
\qquad
p_{\max}=p_b.
\end{equation}
The trained Marketplace delegate represents the buyer, so its terminal reward is $r_{\mathrm{buyer}}$. A deal at the seller's reservation price captures all available surplus for the principal and scores $1$, while a deal at the buyer's reservation price captures none and scores $0$. A deal above the buyer's reservation price also receives zero buyer value. Thus, Craigslist and Marketplace use the same normalized surplus-sharing reward and differ only in how the endpoints of the price corridor are specified: Craigslist uses the task-defined target and listing prices, while Marketplace uses the parties' private reservation prices.

\paragraph{Preference-based reward for slot coordination.}
In Calendar, the agent receives its principal's value function
$v(t)\in[0,1]$ over candidate time slots~\citep{srbench2026}. The requestor has a separate value function over the same slots, constructed in opposition to the principal's preferences. Let $\mathcal{Z}$ denote the zone of possible agreement, defined as the set of time slots that are mutually free on both calendars.

We define the principal-best and counterparty-best feasible values as
\begin{equation}
v_{\max}
=
\max_{t\in\mathcal{Z}}v(t),
\qquad
v_{\min}
=
\min_{t\in\mathcal{Z}}v(t).
\end{equation}
For a meeting scheduled at
$t_{\mathrm{deal}}\in\mathcal{Z}$, the Calendar reward is
\begin{equation}
r_{\mathrm{calendar}}
=
\frac{
v(t_{\mathrm{deal}})-v_{\min}
}{
v_{\max}-v_{\min}
}.
\label{eq:calendar-reward}
\end{equation}
By construction, the feasible slots have different preference scores, so
$v_{\max}>v_{\min}$. The principal's most preferred mutually feasible slot scores $1$, while the slot most favorable to the requestor scores $0$.

This reward distinguishes agreement quality from task completion. Scheduling any mutually feasible meeting completes the coordination task, while the reward measures whether the selected slot actually advances the principal's preferences.

\paragraph{Non-agreement outcomes and common scale.}
Timeouts and aborted interactions receive zero reward. Where an environment provides an explicit walk-away action, we retain its task-specific outside-option payoff. Calendar and Marketplace assign zero outcome value when no agreement is reached.

Across all six environments, the resulting training signal lies in $[0,1]$. The environment-specific constructions preserve the relevant utility structure: fair and efficient allocation, division of price surplus, or preference-sensitive coordination, while the shared range makes reward statistics comparable and provides a consistent interface for the multi-domain consolidation methods in Section~\ref{sec:training}.

\section{SocialRL Training and Unification}
\label{sec:training}

Our study proceeds in two stages. In Stage~1, we train a domain-specialized policy for each of the six environments. PPO directly from the base model is sufficient for Deal-or-No-Deal, CaSiNo, Job Interview, and Calendar, while Craigslist and Marketplace use an SFT warm start before PPO. We then evaluate every specialist on all six environments, revealing substantial and highly asymmetric cross-environment transfer that depends on interaction structure (\S\ref{sec:training:transfer}). In Stage~2, we consolidate the specialists into a single multi-domain policy through two complementary approaches: transfer-aware cascade RL, which uses the observed transfer structure to prioritize final performance, and multi-teacher on-policy distillation (MOPD), which transfers most of the specialists' advantage with substantially less additional training (\S\ref{sec:training:merge}). As a complementary study, we introduce explicit theory-of-mind supervision through \textsc{Infer} $\rightarrow$ \textsc{Act} $\rightarrow$ \textsc{Anticipate} and test whether direct supervision of opponent modeling improves social reasoning and cross-environment generalization (\S\ref{sec:training:tom}). All experiments use Qwen3-4B-Instruct-2507 as the base policy, providing a controlled test of whether targeted post-training can induce broad strategic capabilities in a compact model. Across our six-environment evaluation, the resulting specialists and unified policy achieve aggregate performance in the range of GPT-4.1, GPT-5.1, and GPT-5.2.

\subsection{Stage 1: In-Domain Specialist Training}
\label{sec:training:indomain}

We first train a specialized 4B policy for each of the six environments. For Deal-or-No-Deal, CaSiNo, Job Interview, and Calendar, we optimize the base model directly with PPO using the terminal outcome rewards defined in \S\ref{sec:datasets:reward}. For the two price-negotiation environments, Craigslist and Marketplace, we initialize the policy with supervised fine-tuning (SFT) before applying PPO. This warm start places prerequisite bargaining behaviors within the policy's support, improving exploration; PPO then optimizes these behaviors for the environment's terminal utility. All rollouts use the decoupled infrastructure described in \S\ref{sec:training-infrastructure}. Agents interact with their environments through the standard harness, while the rollout proxy records the model inputs, outputs, rollout log probabilities, and terminal rewards required for training. This separation allows the same PPO implementation to be used across environments with different interaction structures and agent implementations.

\paragraph{PPO configuration.}
Unless otherwise stated, all PPO runs use a separate Qwen3-1.7B critic. We first warm up the critic for $30$ optimization steps while keeping the 4B actor frozen, and then jointly update the actor and critic. We use learning rate of $2\times10^{-6}$ for the actor with batch size 144, mini batch size 36, max gradient norm 5, and learning rate of $1\times10^{-5}$ for the critic. Max training steps is set to be 200. Under this configuration, the critic typically reaches an explained variance of approximately $0.8$ before actor being trained, indicating that the value model captures most of the variation in episodic returns and provides a reliable baseline for PPO updates.

\begin{table}[t]
\centering
\small
\setlength{\tabcolsep}{4.5pt}
\begin{tabular}{@{}lccccccc@{}}
\toprule
& \textbf{DnD} & \textbf{CaSiNo} & \textbf{Craigslist} &
\textbf{Job Int.} & \textbf{Calendar} & \textbf{Mktplace} & \textbf{Avg.} \\
\midrule
Base 4B
    & $0.583\pm0.001$
    & $0.476\pm0.014$
    & $0.318\pm0.012$
    & $0.479\pm0.032$
    & $0.301\pm0.017$
    & $0.174\pm0.026$
    & 0.389 \\

\textbf{Domain-trained 4B}
    & $0.656\pm0.008$
    & $0.503\pm0.007$
    & $0.583\pm0.008$
    & $0.594\pm0.007$
    & $0.540\pm0.017$
    & $0.838\pm0.013$
    & \textbf{0.619} \\
\midrule
GPT-4.1
    & 0.653 & 0.491 & 0.540 & 0.588 & 0.673 & 0.804 & 0.625 \\
GPT-5.1
    & 0.671 & 0.499 & 0.607 & 0.596 & 0.573 & 0.767 & 0.619 \\
GPT-5.2
    & 0.663 & 0.488 & 0.577 & 0.579 & 0.643 & 0.727 & 0.613 \\
GPT-5.5
    & 0.665 & 0.559 & 0.746 & 0.590 & 0.702 & 0.985 & 0.708 \\
\bottomrule
\end{tabular}
\caption{\textbf{In-domain specialist training.} Base and domain-trained 4B scores are reported as mean $\pm$ standard deviation across repeated evaluations; GPT baselines are reported as mean scores. Each domain-trained checkpoint is evaluated on its training environment. The six 4B specialists average 0.619 utility, placing their aggregate performance in the range of GPT-4.1, GPT-5.1, and GPT-5.2.}
\label{tab:indomain}
\end{table}

As shown in Table \ref{tab:indomain}, in-domain post-training produces strong specialists across all six interaction structures \footnote{We ran evaluation for 5 times to compute the mean and variance throughout the paper.}. Averaged across domains, the specialized 4B policies achieve 0.615 utility, comparable to GPT-4.1 (0.625), GPT-5.1 (0.619), and GPT-5.2 (0.613). At the environment level, the 4B specialists fall within or above the GPT-4.1/5.1/5.2 range on DnD, CaSiNo, Craigslist, and Marketplace, while remaining close on Job Interview and Calendar. We next ask whether these domain-specific capabilities transfer across interaction structures.

\subsection{Cross-Environment Transfer}
\label{sec:training:transfer}

We next evaluate every domain-trained policy on all six environments, producing the transfer matrix in Table~\ref{tab:transfer}. The diagonal measures in-domain specialization, while each off-diagonal entry measures how training on one \emph{donor} environment changes performance on another. We summarize each donor by its \emph{out-transfer}, the mean change relative to the base model over the other five environments.

\begin{table}[t]
\centering
\small
\setlength{\tabcolsep}{3.7pt}
\begin{tabular}{@{}lccccccc@{}}
\toprule
\textbf{Donor}
& \textbf{DnD}
& \textbf{CaSiNo}
& \textbf{Craigslist}
& \textbf{Job Int.}
& \textbf{Calendar}
& \textbf{Mktplace}
& \textbf{Out-tr.} \\
\midrule
Baseline
& $0.583{\pm}.001$
& $0.476{\pm}.014$
& $0.318{\pm}.012$
& $0.479{\pm}.032$
& $0.301{\pm}.012$
& $0.174{\pm}.003$
& --- \\
\midrule
Deal-or-No-Deal
& $\mathbf{0.656{\pm}.008}^{\star}$
& $0.491{\pm}.012$
& $0.282{\pm}.028$
& $0.481{\pm}.016$
& $0.370{\pm}.009$
& $0.150{\pm}.003$
& $+0.005$ \\

CaSiNo
& $0.597{\pm}.019$
& $\mathbf{0.503{\pm}.007}^{\star}$
& $0.309{\pm}.004$
& $0.463{\pm}.016$
& $0.331{\pm}.010$
& $0.168{\pm}.022$
& $+0.003$ \\

Craigslist$^{\dagger}$
& $0.564{\pm}.023$
& $0.494{\pm}.007$
& $\mathbf{0.583{\pm}.008}^{\star}$
& $0.447{\pm}.011$
& $0.266{\pm}.023$
& $0.491{\pm}.007$
& $\mathbf{+0.050}$ \\

Job Interview
& $0.610{\pm}.017$
& $0.511{\pm}.004$
& $0.328{\pm}.010$
& $\mathbf{0.594{\pm}.007}^{\star}$
& $0.373{\pm}.022$
& $0.161{\pm}.011$
& $+0.026$ \\

Calendar
& $0.607{\pm}.010$
& $0.474{\pm}.007$
& $0.167{\pm}.013$
& $0.311{\pm}.013$
& $\mathbf{0.540{\pm}.017}^{\star}$
& $0.172{\pm}.008$
& $-0.060$ \\

Marketplace
& $0.558{\pm}.010$
& $0.464{\pm}.007$
& $0.502{\pm}.015$
& $0.371{\pm}.010$
& $0.268{\pm}.015$
& $\mathbf{0.838{\pm}.013}^{\star}$
& $+0.001$ \\
\midrule
GPT-4.1
& 0.653 & 0.491 & 0.540 & 0.588 & 0.673 & 0.804 & --- \\
GPT-5.1
& 0.671 & 0.499 & 0.607 & 0.596 & 0.573 & 0.767 & --- \\
GPT-5.2
& 0.663 & 0.488 & 0.577 & 0.579 & 0.643 & 0.727 & --- \\
GPT-5.5
& 0.665 & 0.559 & 0.746 & 0.590 & 0.702 & 0.985 & --- \\
\bottomrule
\end{tabular}
\caption{\textbf{Cross-environment transfer.}
Each row evaluates a policy trained on one donor environment across all six domains ($\star$ denotes the training domain). Out-transfer is the mean change from the base 4B policy over the five off-domain environments. $^{\dagger}$Craigslist uses SFT+PPO. Reported uncertainties are standard deviations across three SGLang benchmark runs where available.}
\label{tab:transfer}
\end{table}

Transfer depends strongly on the donor-recipient configuration. The clearest pattern is transfer between environments with similar interaction structure. Craigslist and Marketplace, both price-negotiation environments, exhibit the two largest cross-domain gains: training on Craigslist raises Marketplace from $0.174$ to $0.491$ ($+0.317$), while training on Marketplace raises Craigslist from $0.318$ to $0.502$ ($+0.184$). The allocation pair DnD and CaSiNo also transfers positively in both directions, although more modestly. Some environments transfer more broadly. Job Interview is the strongest general-purpose donor after Craigslist, with an average out-transfer of $+0.026$ and improvements on DnD, CaSiNo, Craigslist, and Calendar. In contrast, a policy can become very strong in-domain without becoming a useful donor: Marketplace reaches $0.838$ on its own task while its average effect on the other five environments is approximately neutral. Calendar is the clearest negative donor in some domains, with out-transfer of $-0.060$, including large drops on Craigslist ($0.318\!\rightarrow\!0.167$) and Job Interview ($0.479\!\rightarrow\!0.311$).

These results show that transfer is strongly asymmetric and structured: what matters is not simply how strong a specialist is, but which capability it learns and which environment receives it. This observation is central to unification. In the next stage, we exploit the measured transfer structure to choose the ordering of environments for cascade RL, and compare this transfer-aware ordering against random and anti-transfer orderings.

\subsection{Stage 2: Consolidating Specialists into a Unified Model}
\label{sec:training:merge}

Stage~1 produces a strong specialist for each environment. Our next goal is to consolidate these capabilities into a single policy that performs well across all six interaction structures. This is non-trivial because training effects are coupled across domains: as shown by the transfer matrix in Table~\ref{tab:transfer}, optimizing one environment can improve, preserve, or interfere with performance on another. Sequential training can therefore both exploit positive transfer and induce catastrophic forgetting. We study two complementary approaches to unification, targeting different points in the performance-efficiency tradeoff.

\paragraph{Two routes to unification.}
Our first approach, \emph{cascade RL}~\citep{wang2025nemotron, du2024cascading}, continues reinforcement learning sequentially across environments. Because each stage directly optimizes environment reward, cascade RL can exploit positive transfer during training and may even improve previously learned capabilities beyond the corresponding single-domain specialists. Its cost is substantial: each environment requires another full RL stage, effectively relearning the capabilities rather than directly reusing the trained specialists. Moreover, the final policy depends strongly on training order, since beneficial and destructive transfer are highly asymmetric. We therefore explore the construction of a transfer-aware curriculum based on the transfer matrix from Section~\ref{sec:training:transfer}.

Our second approach, MOPD~\citep{ma2026mopd, yang2024multi, yang2026learning, fu2026poly}, reuses the Stage~1 specialists directly as teachers and distills their capabilities into a shared student. This is substantially more efficient: fewer than $100$ additional optimization steps recover most of the specialists' advantage. The main challenge shifts from environment ordering to allocation of the distillation budget: different teachers provide very different amounts of additional capability over the student, and some have already been nearly matched through cross-domain transfer. We address this with a gap-closed curriculum that focuses sampling on domains whose teacher advantage remains unabsorbed.

\paragraph{Shared initialization.}
Both consolidation methods start from the same Craigslist-SFT checkpoint. This initialization is necessary because the base 4B model rarely explores the anchoring behavior required for successful Craigslist negotiation: as observed in Stage~1, direct PPO fails to improve the base policy on Craigslist. We find the same limitation for OPD from the base model also fails to acquire the Craigslist capability when trained on Craigslist alone. Supervised distillation first places this prerequisite behavior within the student's support, after which either RL or OPD can optimize and combine it with capabilities from the other environments. Using the same initialization for both methods also makes their comparison controlled.

Cascade RL and MOPD thus provide complementary approaches to the same consolidation problem. Cascade RL prioritizes final performance by continuing reward optimization and exploiting cross-environment transfer, while MOPD prioritizes efficiency by directly transferring the capabilities already learned by the domain specialists.

\subsubsection{Stage 2A: Transfer-Aware Unification with Cascade RL}
\label{sec:training:cascade}

\paragraph{Training principle.}
Cascade RL consolidates capabilities by continuing reinforcement learning sequentially across environments. A single policy is carried through the cascade: at each stage, PPO optimizes the policy on one environment using the same outcome reward as in Stage~1, and the selected checkpoint initializes the next stage. We periodically evaluate checkpoints on all environments encountered so far and select the one with the highest average utility, balancing progress on the current domain against retention of previously acquired capabilities.

The central design choice is therefore the \emph{order} of environments. The transfer matrix in Table~\ref{tab:transfer} shows that cross-environment effects are strongly directional: training on one domain may improve another, leave it largely unchanged, or substantially degrade it. Sequential RL is consequently path-dependent. We use the observed transfer structure to construct a curriculum that places beneficially interacting domains together and schedules destructive donors before the capabilities they would otherwise overwrite.

\paragraph{Transfer-aware ordering.}
Three patterns in Table~\ref{tab:transfer} guide the curriculum. First, Craigslist and Marketplace form the strongest transfer pair in the suite: Craigslist training improves Marketplace by $+0.317$, while Marketplace improves Craigslist by $+0.184$. We therefore place them consecutively, with Craigslist first to exploit the stronger transfer direction. Second, Calendar is a destructive donor for Craigslist and Job Interview: reducing Craigslist by $0.151$ and Job Interview by $0.168$, and is the only domain with negative average out-transfer. We therefore place Calendar at the beginning of the cascade, before these bargaining capabilities are acquired. Third, Job Interview exhibits strongly asymmetric transfer. It transfers positively to several other domains, yet its own performance is poorly preserved after training elsewhere. We therefore place it last, allowing the final stage to recover Job Interview capability without exposing it to subsequent interference.

These considerations yield the \textit{transfer-aware} curriculum
\begin{equation}
\text{Calendar}
\rightarrow
\text{CaSiNo}
\rightarrow
\text{DnD}
\rightarrow
\text{Craigslist}
\rightarrow
\text{Marketplace}
\rightarrow
\text{Job Interview}.
\label{eq:cascade-order}
\end{equation}

\paragraph{CascadeRL configuration.}
All cascade experiments start from the same Craigslist-SFT initialization and use the same PPO configuration and per-stage training budget. Each stage is an independent PPO with max 200 steps, and we evaluate each checkpoint every 20 training steps on a small validation dataset.

\paragraph{Ordering controls.}
To test whether the measured transfer structure provides a useful curriculum signal, we compare the transfer-aware ordering with alternative sequences while holding initialization and training budget fixed. Our random-order control uses
\begin{equation}
\text{Craigslist}
\rightarrow
\text{Job Interview}
\rightarrow
\text{DnD}
\rightarrow
\text{CaSiNo}
\rightarrow
\text{Calendar}
\rightarrow
\text{Marketplace}.
\label{eq:cascade-random}
\end{equation}
We additionally construct an \emph{anti-transfer} curriculum that deliberately reverses the main ordering principles:
\begin{equation}
\text{Job Interview}
\rightarrow
\text{Marketplace}
\rightarrow
\text{Craigslist}
\rightarrow
\text{DnD}
\rightarrow
\text{CaSiNo}
\rightarrow
\text{Calendar}.
\label{eq:cascade-anti}
\end{equation}
This ordering places the difficult-to-preserve Job Interview capability first and the destructive Calendar stage last. 

\paragraph{Unified-model performance.}
Table~\ref{tab:cascade-frontier} reports the final six-environment performance of the transfer-aware, random-order, and anti-transfer cascades, together with the GPT baselines. The transfer-aware cascade reaches an Avg-6 of $0.627\pm0.004$, placing the unified 4B policy in the same aggregate performance range as GPT-4.1 ($0.625$), GPT-5.1 ($0.619$), and GPT-5.2 ($0.613$). In contrast, the random-order cascade reaches only $0.584\pm0.008$ and the anti-transfer cascade reaches only $0.562\pm0.010$ under the same overall training setup.

The aggregate result masks substantial cross-domain interaction. Calendar is the clearest example of positive accumulation: although it is trained at the very beginning of the cascade, the final policy reaches $0.742$, well above its single-domain specialist score of $0.540$ and above every evaluated GPT baseline. Marketplace is also strongly retained at $0.803$, essentially matching GPT-4.1 at $0.804$, while Craigslist reaches $0.580$, above GPT-4.1 and GPT-5.2. Thus, sequential RL can do more than preserve earlier specialists: subsequent stages can reinforce capabilities acquired earlier through positive cross-environment transfer.

\paragraph{Stage-by-stage transfer dynamics.}
The final benchmark does not reveal how capabilities evolve as the policy moves through the cascade. Table~\ref{tab:cascade-stages} traces the checkpoint selected at the end of each stage, evaluated on the environments included in checkpoint selection at that point on a small validation dataset. Because every cascade starts from the Craigslist-SFT initialization, Craigslist is tracked from the first stage even though its PPO stage occurs fourth. These values are the within-cascade evaluations used for checkpoint selection; the final repeated benchmark is reported separately in Table~\ref{tab:cascade-frontier}.

\begin{table*}[!ht]
\centering
\small
\setlength{\tabcolsep}{4.0pt}
\begin{tabular}{@{}lcccccccc@{}}
\toprule
\textbf{Stage completed}
& \textbf{Step}
& \textbf{DnD}
& \textbf{CaSiNo}
& \textbf{Craigslist}
& \textbf{Job Int.}
& \textbf{Calendar}
& \textbf{Mktplace}
& \textbf{Mean seen} \\
\midrule
Calendar
& 100
& ---
& ---
& 0.614
& ---
& \textbf{0.621}
& ---
& 0.617 \\

CaSiNo
& 40
& ---
& \textbf{0.504}
& 0.517
& ---
& 0.733
& ---
& 0.585 \\

Deal-or-No-Deal
& 100
& \textbf{0.618}
& 0.498
& 0.489
& ---
& 0.758
& ---
& 0.591 \\

Craigslist
& 40
& 0.628
& 0.477
& \textbf{0.544}
& ---
& 0.729
& ---
& 0.595 \\

Marketplace
& 140
& 0.628
& 0.483
& 0.564
& ---
& 0.704
& \textbf{0.842}
& 0.644 \\

Job Interview
& 60
& 0.620
& 0.513
& 0.570
& \textbf{0.529}
& 0.692
& 0.823
& 0.625 \\
\bottomrule
\end{tabular}
\caption{\textbf{Stage-by-stage evolution of the transfer-aware cascade.} Each row reports the checkpoint selected after optimizing the environment in the first column; bold denotes the environment optimized at that stage. Craigslist is evaluated from the beginning because the cascade is initialized from its SFT checkpoint. Dashes denote environments not yet included in checkpoint selection. Mean seen averages the environments tracked at that stage and is therefore not directly comparable across rows as the set of environments expands. These are within-cascade checkpoint-selection evaluations; Table~\ref{tab:cascade-frontier} reports the final repeated six-environment benchmark.}
\label{tab:cascade-stages}
\end{table*}

The stagewise trajectory shows that cross-environment transfer occurs during the cascade itself. Calendar reaches $0.621$ after its own stage, then improves to $0.733$ after CaSiNo and $0.758$ after DnD despite receiving no additional Calendar-specific updates. Later bargaining stages introduce some forgetting, but Calendar remains strong at $0.692$ after the final Job Interview stage. Its high final performance therefore reflects capability accumulated across multiple stages, rather than retention of the original Calendar checkpoint alone.

The Craigslist-Marketplace pair exhibits a second form of transfer. Craigslist performance initially falls during the allocation stages ($0.614\rightarrow0.517\rightarrow0.489$), recovers to $0.544$ when Craigslist is optimized directly, and then rises further to $0.564$ after Marketplace training. At the same checkpoint, Marketplace reaches $0.842$ while DnD and Calendar remain at $0.628$ and $0.704$, respectively. The final Job Interview stage preserves most of these gains: Marketplace remains at $0.823$ and Craigslist rises to $0.570$, hile also raising CaSiNo from $0.483$ to $0.513$.

These dynamics are neither monotonic nor uniformly positive. Earlier capabilities can temporarily degrade, as Craigslist does during the allocation stages, and Calendar declines from its intermediate peak as later domains are introduced. The role of the transfer-aware curriculum is therefore to order these interactions so that destructive stages occur early and later stages can repair or reinforce related capabilities. Checkpoint selection on the environments seen so far further limits forgetting at each transition.

\begin{table}[!ht]
\centering
\small
\setlength{\tabcolsep}{3.0pt}
\begin{tabular}{@{}lccc@{}}
\toprule
\textbf{Environment}
& \textbf{Transfer-aware}
& \textbf{Random}
& \textbf{Anti-transfer} \\
\midrule
Deal-or-No-Deal
& $0.617\pm0.008$
& $0.604\pm0.012$
& $0.622\pm0.018$ \\

CaSiNo
& $0.482\pm0.019$
& $0.465\pm0.016$
& $0.443\pm0.023$ \\

Craigslist
& $0.580\pm0.007$
& $\mathbf{0.637\pm0.022}$
& $0.497\pm0.018$ \\

Job Interview
& $0.538\pm0.012$
& $0.478\pm0.030$
& $0.471\pm0.009$ \\

Calendar
& $\mathbf{0.742\pm0.037}$
& $0.541\pm0.021$
& $0.641\pm0.011$ \\

Marketplace
& $0.803\pm0.008$
& $0.783\pm0.024$
& $0.698\pm0.022$ \\
\midrule
\textbf{Avg-6}
& $\mathbf{0.627\pm0.004}$
& $0.578\pm0.008$
& $0.562\pm0.010$ \\
\bottomrule
\end{tabular}
\caption{\textbf{Cascade RL is sensitive to environment ordering.} The transfer-aware curriculum follows Calendar$\rightarrow$CaSiNo$\rightarrow$DnD$\rightarrow$Craigslist $\rightarrow$Marketplace$\rightarrow$Job Interview; the random-order control follows Eq.~\ref{eq:cascade-random}; and the anti-transfer control follows Eq.~\ref{eq:cascade-anti}. Each final evaluation uses $10$ scenarios $\times$ $3$ opponents $\times$ $5$ trials ($150$ games per environment). Transfer-aware scores are mean $\pm$ standard deviation over two independent evaluations, and random-order scores are mean $\pm$ standard deviation over repeated evaluations. Anti-transfer performs way worse than both Transfer-aware as well as Random-order. The transfer-aware unified 4B policy achieves an Avg-6 of $0.627$, in the performance range of GPT-4.1, GPT-5.1, and GPT-5.2.}
\label{tab:cascade-frontier}
\end{table}

\paragraph{Comparison with other curriculums.}
The comparison with random ordering and anti-optimal ordering shows that this accumulation depends strongly on curriculum. The transfer-aware cascade improves Avg-6 from $0.584$ to $0.627$, with particularly large gains on Calendar ($0.541\rightarrow0.742$), Job Interview ($0.478\rightarrow0.538$), and Marketplace ($0.783\rightarrow0.803$). The random ordering performs better on Craigslist in isolation ($0.637$ vs.\ $0.580$), but fails to preserve comparable performance across the full suite. This is precisely the objective of the transfer-aware curriculum: optimize the final multi-domain policy rather than any single stage.

These results support the use of the transfer matrix as a reasonable curriculum-design signal. The same environments, trained with the same initialization and learning procedure, can lead to substantially different unified policies depending on their order. Cascade RL can therefore achieve strong multi-domain performance by exploiting the directionality of cross-environment transfer, at the cost of running a full sequence of reinforcement-learning stages.

\subsubsection{Stage 2B: Efficient Unification with MOPD}
\label{sec:training:mopd}

\paragraph{Training principle.}
Cascade RL continues optimizing environment rewards, but requires a full RL stage for each additional domain. MOPD instead treats the Stage-1 specialists as teachers and directly transfers their capabilities into a common student. At each rollout, we sample an environment $e$ and pair it with its corresponding specialist. The student generates trajectories on-policy, and its responses are trained toward the teacher using a reverse-KL distillation objective. This construction preserves the original interaction stack: the environment and agent harness execute unchanged, while the rollout proxy records the student trajectory and queries the corresponding teacher for distillation. Changing from one teacher to six therefore requires no modification to the environments or agent implementations.

\paragraph{MOPD configuration.}
The MOPD runs use a reverse-KL objective: the student generates every rollout on-policy, and the corresponding domain specialist supplies token-level teacher logits for the generated responses. We truncate the teacher distribution to its top $32$ logits, use a distillation temperature of $1.0$, and apply the loss uniformly over all response tokens without position weighting. We optimize with a constant learning rate of $10^{-5}$ without warmup.

\paragraph{Initialization and teacher selection.}
We initialize the unified student from the Craigslist-SFT checkpoint used throughout Stage~2. We also exclude teachers that offer essentially no additional headroom over this initialization. CaSiNo is the clearest case: the Craigslist-SFT student already scores $0.502$ on CaSiNo, nearly matching the domain specialist at $0.503$. Distilling from such a teacher provides little additional signal while potentially introducing optimization noise. We therefore omit CaSiNo from the MOPD training mixture and retain it only for evaluation. The transfer results in Table~\ref{tab:transfer} further suggest that CaSiNo performance can be maintained through capabilities learned from other domains.

\paragraph{Gap-closed curriculum.}
The remaining teachers still differ substantially in how much capability the student has left to absorb. Uniform sampling gives equal training budget to domains that are nearly solved and domains with large remaining gaps. We instead adapt the sampling distribution to the fraction of each teacher's advantage that remains unclosed: Let $\mathcal{E}$ denote the candidate training environments. For each $e\in\mathcal{E}$, let $B_e$ be the student's pre-distillation score, $T_e$ its specialist teacher's score, and $\bar{s}_e^{(t)}$ the student's rolling mean terminal reward over the most recent $W=500$ training games. Before any observations are available, we initialize $\bar{s}_e^{(t)}=B_e$.

We define
\begin{equation}
g_e = T_e-B_e,
\qquad
\rho_e^{(t)}
=
\frac{\bar{s}_e^{(t)}-B_e}{g_e},
\label{eq:mopd-gap}
\end{equation}
where $g_e$ is the teacher-student headroom and $\rho_e^{(t)}$ is the fraction already closed. The unnormalized sampling weight is
\begin{equation}
w_e^{(t)}
=
\begin{cases}
\lambda,
& g_e < \delta, \\[4pt]
\max\!\left(0,1-\rho_e^{(t)}\right)+\lambda,
& \text{otherwise},
\end{cases}
\label{eq:mopd-weight}
\end{equation}
with floor $\lambda=0.05$ and minimum-gap threshold $\delta=0.03$. We normalize
\begin{equation}
p_e^{(t)}
=
\frac{w_e^{(t)}}{\sum_{e'} w_{e'}^{(t)}},
\end{equation}
and band-clamp the resulting probabilities to
\begin{equation}
p_e^{(t)}\in[p_{\min},p_{\max}],
\end{equation}
iteratively renormalizing the remaining mass. $p_{\min},p_{\max}$ are set to be $[0.10,0.40]$ in our experiments. The normalization by each domain's own gap makes progress comparable across environments with different reward ranges. At initialization, $\rho_e=0$ for every learnable environment. As the student approaches its teacher, the corresponding sampling mass decreases; once it matches or surpasses the teacher, the environment receives only the floor weight. The minimum-gap guard prevents near-zero teacher-student gaps from turning reward noise into extreme sampling weights, while the probability band prevents any remaining environment from dominating or disappearing from training.

\paragraph{Evaluation metrics.}
We evaluate every unified checkpoint on all six environments and report selected checkpoints as mean $\pm$ standard deviation over five independent evaluation runs. We use three complementary aggregate metrics.

The first is the unweighted mean utility which directly measures overall model performance:
\begin{equation}
\mathrm{Avg6}(s)
=
\frac{1}{6}\sum_{e} s_e ,
\end{equation}

To measure the objective specific to specialist consolidation, we additionally compute the fraction of each teacher's advantage absorbed by the student:
\begin{equation}
C_e(s)
=
\frac{s_e-B_e}{T_e-B_e}.
\label{eq:gap-closure}
\end{equation}
A value of $1$ means that the unified student matches the specialist teacher, and values above $1$ indicate that it surpasses the specialist. CaSiNo is excluded because its teacher-initialization gap is only $0.001$.

Finally, we report clipped gap closure where $\mathcal{E}'$ excludes CaSiNo.
\begin{equation}
\bar{C}_{\mathrm{clip}}
=
\frac{1}{|\mathcal{E}'|}
\sum_{e\in\mathcal{E}'}
\operatorname{clip}\!\left(C_e,0,1\right),
\end{equation} 
Clipping prevents over-performance on one environment from compensating for incomplete transfer on another. Avg-6 thus measures absolute capability, while the two gap-closure metrics measure how completely the specialist capabilities have been consolidated.

\paragraph{Curriculum ablations.}
Table~\ref{tab:mopd} compares adaptive and uniform environment sampling, with and without the saturated CaSiNo teacher. The gap-closed curriculum without CaSiNo performs best under all three metrics. After only $60$ optimization steps, it reaches an Avg-6 of $0.597\pm0.015$, with $92.6\%$ mean gap closure and $84.5\%$ clipped gap closure. Removing CaSiNo helps under both uniform and adaptive sampling, confirming that allocating distillation budget to a teacher with essentially zero headroom is counterproductive.

\begin{table}[t]
\centering
\small
\begin{tabular}{@{}lcccc@{}}
\toprule
\textbf{Configuration}
& \textbf{Avg-6 $\uparrow$}
& \textbf{Gap closure $\uparrow$}
& \textbf{Clipped gap closure $\uparrow$} \\
\midrule
\textbf{Gap-closed, w/o CaSiNo}
& $\mathbf{0.597\pm0.015}$
& $\mathbf{92.6\pm11.3}$\%
& $\mathbf{84.5\pm6.3}$\% \\
Equal mix, w/o CaSiNo
& $0.594\pm0.020$
& $84.4\pm15.6$\%
& $79.7\pm12.6$\% \\
Equal mix, w/ CaSiNo
& $0.595\pm0.015$
& $82.2\pm9.2$\%
& $77.6\pm5.8$\% \\
Gap-closed, w/ CaSiNo
& $0.578\pm0.012$
& $71.3\pm8.7$\%
& $70.7\pm7.9$\% \\
\midrule
Student initialization
& 0.460
& ---
& --- \\
Teacher average
& 0.619
& ---
& --- \\
\bottomrule
\end{tabular}
\caption{\textbf{MOPD.}
All configurations use reverse-KL distillation without position weighting and a learning rate of $10^{-5}$. Scores are mean $\pm$ standard deviation over five independent evaluation runs. The gap-closed curriculum without the saturated CaSiNo teacher performs best under all three aggregate metrics.}
\label{tab:mopd}
\end{table}

\paragraph{Per-environment consolidation.}
Table~\ref{tab:mopd-per-env} shows how the best unified checkpoint compares with its initialization and specialist teachers. The unified model recovers $85\%$ of the Craigslist teacher's advantage, $69\%$ on Job Interview, and $83\%$ on Marketplace. It also surpasses the corresponding specialist on DnD ($0.664$ vs.\ $0.656$) and Calendar ($0.564$ vs.\ $0.540$), yielding gap closures above $100\%$. Thus, MOPD can combine specialist capabilities without constraining the unified model to interpolate below each teacher.

\begin{table}[!ht]
\centering
\small
\setlength{\tabcolsep}{5pt}
\begin{tabular}{@{}lcccc@{}}
\toprule
\textbf{Environment}
& \textbf{Unified 4B}
& \textbf{Initialization}
& \textbf{Teacher}
& \textbf{Gap closed} \\
\midrule
Deal-or-No-Deal
& $\mathbf{0.664 \pm 0.025}$
& 0.586
& $0.656\pm 0.008$
& $\mathbf{111\%}$ \\

Craigslist
& $0.556 \pm 0.019$
& 0.397
& $0.583\pm 0.007$
& $85\%$ \\

CaSiNo$^{\dagger}$
& $0.478 \pm 0.019$
& 0.502
& $0.503\pm 0.008$
& --- \\

Job Interview
& $0.551 \pm 0.023$
& 0.456
& $0.594\pm 0.007$
& $69\%$ \\

Calendar
& $\mathbf{0.564 \pm 0.050}$
& 0.375
& $0.540\pm 0.017$
& $\mathbf{115\%}$ \\

Marketplace
& $0.771 \pm 0.028$
& 0.444
& $0.838\pm 0.013$
& $83\%$ \\
\midrule
\textbf{Avg-6}
& $\mathbf{0.597 \pm 0.015}$
& 0.460
& 0.619
& $\mathbf{92.6\%}^{*}$ \\
\bottomrule
\end{tabular}
\caption{\textbf{Per-environment performance of the best MOPD model.}
We report the gap-closed curriculum without CaSiNo at step 60. Unified-model scores are mean $\pm$ standard deviation over five independent evaluation runs. Gap closure measures the fraction of each specialist's advantage over the initialization recovered by the unified model. $^{\dagger}$CaSiNo is excluded from MOPD training and gap-closure aggregation because its teacher--initialization gap is only $0.001$. $^{*}$Mean gap closure over the remaining five environments.}
\label{tab:mopd-per-env}
\end{table}

Overall, MOPD provides a substantially cheaper consolidation route than sequential RL: most of the specialist advantage is transferred in fewer than $100$ additional optimization steps. Together, the two Stage-2 methods expose a performance-efficiency tradeoff: cascade RL uses continued reward optimization and transfer-aware ordering to pursue a stronger final policy, while MOPD directly compresses the specialists into a unified model with a much smaller additional training budget.

\subsection{Explicit Theory-of-Mind Supervision}
\label{sec:training:tom}

The training recipe so far improves negotiation through outcome-based reinforcement learning: the model is rewarded for reaching favorable agreements, while any reasoning about the counterpart is learned only indirectly. Yet effective negotiation depends naturally on understanding the other party~\citep{hwang2026infusing, mu2026adaptive, kostka2025evaluating, xiao2025towards}. An agent must infer what the counterpart values, identify where their preferences overlap or conflict, and anticipate how the counterpart is likely to respond to a proposed action. We therefore study a complementary approach that supervises these intermediate reasoning capabilities explicitly, asking whether stronger \emph{theory of mind} (ToM) can translate into stronger negotiation.

\paragraph{Theory-of-mind capabilities.}
We operationalize theory of mind through two capabilities that are directly relevant to strategic interaction. \emph{Preference inference} (\textsc{Infer}) measures whether the agent can recover the counterpart's latent preferences or private state from the interaction history. \emph{Next-action prediction} (\textsc{Anticipate}) measures whether the agent can predict how the counterpart will respond to the agent's planned move. Together with the agent's own action (\textsc{Act}), these form the reasoning scaffold
\begin{equation}
\textsc{Infer} \;\rightarrow\; \textsc{Act} \;\rightarrow\; \textsc{Anticipate}.
\end{equation}
For each decision, the scaffold records the agent's current belief about the counterpart, its chosen action, and its prediction of the counterpart's subsequent reaction. We evaluate \textsc{Infer} by exact-match accuracy against the counterpart's private preferences and \textsc{Anticipate} by the accuracy of the predicted next action.

\paragraph{Does prompting alone elicit theory of mind?}
We first test whether the base 4B model can benefit from the scaffold without additional training. The answer is negative: adding the explicit ToM prompt reduces average negotiation utility from $0.454$ to $0.353$ (Table~\ref{tab:tom}), with the largest drop on Job Interview ($0.476\rightarrow0.221$). Requiring a small model to produce an explicit reasoning structure therefore does not by itself improve strategic behavior.

Direct evaluation of the two ToM capabilities helps explain this result. The base model already performs reasonably well at preference inference, reaching $0.618$ accuracy, but is substantially weaker at next-action prediction, at approximately $54\%$. Thus, the base model can often recover what the counterpart wants while remaining much less capable of anticipating what the counterpart will actually do. Prompting exposes this limitation rather than resolving it.

\paragraph{Training explicit theory of mind.}
We next train these capabilities directly through supervised distillation. GPT-5.2 generates demonstrations containing both the ordinary negotiation action and the explicit \textsc{Infer} and \textsc{Anticipate} reasoning. We refer to SFT on these complete traces as \emph{ExpToM SFT}. The experiment uses no reinforcement learning, allowing us to isolate the effect of explicit reasoning supervision from outcome optimization. We compare ExpToM SFT with standard SFT on action trajectories, under both per-environment training and joint training across the four negotiation environments.

Explicit ToM supervision substantially improves the model's measured reasoning capabilities. Next-action prediction improves by roughly $30$ percentage points, while preference-inference accuracy rises to approximately $0.63$--$0.71$, depending on the training environment. More importantly, these gains translate into stronger negotiation. Per-environment ExpToM SFT reaches an average utility of $0.546$, compared with $0.500$ for ordinary SFT and $0.454$ for the base model. The improvement is consistent across all four environments. The effect also survives multi-domain training: mixed ExpToM reaches $0.525$, compared with $0.495$ for mixed ordinary SFT.

\begin{table}[t]
\centering
\small
\setlength{\tabcolsep}{5pt}
\begin{tabular}{@{}lccccccc@{}}
\toprule
& & & \multicolumn{2}{c}{\textbf{Normal SFT}}
& \multicolumn{2}{c}{\textbf{ExpToM SFT}} & \\
\cmidrule(lr){4-5}\cmidrule(lr){6-7}
\textbf{Environment}
& \textbf{Base 4B}
& \textbf{+ToM prompt}
& \textbf{per-env}
& \textbf{mixed}
& \textbf{per-env}
& \textbf{mixed}
& \textbf{GPT-5.2} \\
\midrule
Deal-or-No-Deal
& 0.589 & 0.575 & 0.580 & 0.587 & 0.596 & \textbf{0.608} & 0.655 \\
CaSiNo
& 0.454 & 0.391 & 0.492 & 0.480 & \textbf{0.512} & 0.498 & 0.475 \\
Craigslist
& 0.295 & 0.225 & 0.445 & 0.409 & \textbf{0.552} & 0.478 & 0.511 \\
Job Interview
& 0.476 & 0.221 & 0.482 & 0.503 & \textbf{0.522} & 0.517 & 0.626 \\
\midrule
Average
& 0.454 & 0.353 & 0.500 & 0.495 & \textbf{0.546} & 0.525 & 0.567 \\
\bottomrule
\end{tabular}
\caption{\textbf{Explicit theory-of-mind supervision improves negotiation.}
Normalized utility for the base model, prompt-only ToM, ordinary SFT, and
explicit ToM SFT. ExpToM distills
\textsc{Infer}$\rightarrow$\textsc{Act}$\rightarrow$\textsc{Anticipate}
traces from GPT-5.2. Prompting the scaffold alone hurts the base 4B, while
training on the same reasoning structure improves over ordinary SFT under
both per-environment and mixed training.}
\label{tab:tom}
\end{table}

\paragraph{Which theory-of-mind capability matters for negotiation?}
Finally, we ask which component of ToM is associated with successful negotiation. Across trained checkpoints, next-action prediction accuracy is positively correlated with negotiation utility, while preference-inference accuracy shows little corresponding relationship. The distinction is visible already in the base model: it has relatively strong preference inference despite weak negotiation performance, while its largest ToM deficit is next-action prediction.

This suggests that identifying the counterpart's preferences is only one part of effective social reasoning. Strategic action additionally requires anticipating how those preferences translate into behavior in response to a particular move. Explicit ToM supervision improves both capabilities, but the improvement in \textsc{Anticipate} is the component most closely associated with better negotiation outcomes.

Taken together, these experiments provide a second route for improving social reasoning alongside outcome-based RL. Reinforcement learning trains strategic behavior from the quality of the final agreement; explicit ToM distillation instead supervises intermediate reasoning about the counterpart. The latter improves negotiation even under pure SFT, showing that part of the capability learned implicitly through interaction can also be transferred directly through structured reasoning supervision.

\section{Qualitative Analysis}
\label{sec:qualitative}

The aggregate results in \S\ref{sec:training} show that post-training substantially improves negotiation utility. We next inspect the interaction trajectories to understand \emph{how} the learned policies differ from the base model. Our current analysis covers the six negotiation environments: for each environment, we compare the base model with the final domain-trained policy; where available, we also examine an SFT-only checkpoint as a diagnostic for separating behaviors introduced by imitation from those selected by reinforcement learning.

We organize the analysis around two questions. First, how does training change the agent's \emph{strategic decisions}: where it anchors, what it concedes, when it rejects, and what value it preserves for its principal? Second, how does training change the \emph{interaction itself}: communication style and action choice? 

\subsection{Strategic Behavior: More Selective Concession}
\label{sec:qualitative:mechanism}

Across the four negotiation environments, the most consistent strategic change is greater selectivity in concession. The base model often moves quickly toward agreement after encountering resistance, even when doing so gives away substantial utility. Post-training makes the policy more willing to maintain a favorable position, reject an unfavorable proposal, and concede on dimensions that are relatively inexpensive. The concrete behavior differs with the structure of each environment.

\paragraph{Craigslist: anchor low and concede gradually.}
Craigslist gives the clearest view of how training reshapes the bargaining trajectory. We normalize buyer offers by the listing-to-target gap, with $0$ denoting the buyer's target price, $1$ the listing price, and negative values offers below the target. Table~\ref{tab:concession} shows the resulting concession curves.

The base model opens essentially at its own target ($+0.04$), with only $3\%$ of openings below target, and then moves rapidly toward the seller. SFT introduces the missing anchoring behavior: the mean opening moves to $-1.21$, and $71\%$ of games begin below the buyer's target. PPO strengthens this strategy further. The SFT$+$PPO policy opens at $-1.48$, remains below target through its third proposal on average, and raises the below-target opening rate to $78\%$.

\begin{table}[t]
\centering
\small
\begin{tabular}{@{}lccccc@{}}
\toprule
\textbf{Stage} & \textbf{Open} & \textbf{2nd} & \textbf{3rd} & \textbf{4th}
& \textbf{\% open $<$ target} \\
\midrule
Untrained & $+0.04$ & $+0.35$ & $+0.41$ & $+0.45$ & 3\% \\
SFT       & $-1.21$ & $-0.30$ & $-0.08$ & $-0.06$ & 71\% \\
SFT$+$PPO & $-1.48$ & $-0.73$ & $-0.22$ & $+0.01$ & 78\% \\
\bottomrule
\end{tabular}
\caption{\textbf{Craigslist buyer concession curves.}
Offer price is normalized by the listing-target gap: $0$ denotes the buyer's target, $1$ the listing price, and negative values offers below target. SFT introduces below-target anchoring; PPO produces a lower initial anchor and a slower subsequent concession trajectory.}
\label{tab:concession}
\end{table}

Matched trajectories illustrate the difference. With a \$275 listing and a \$209 target, the base model opens at \$209 and eventually accepts \$275, obtaining zero utility. The trained policy instead negotiates \$150 $\rightarrow$ \$165, rejects \$250, and eventually accepts \$190, obtaining maximal utility. Training therefore changes both the initial reference point and the willingness to maintain it under pressure.

Craigslist also reveals a division of labor between SFT and PPO in the language used to support these offers. The base buyer frequently reveals its own target price: $52.7\%$ of its messages explicitly mention that private reference point, while only $3.7\%$ invoke external market comparisons. After SFT, target leakage falls to $1.0\%$ and market/comparable references rise to $29.6\%$ of messages. PPO subsequently shifts toward commitment language: explicit ``final offer,'' ``firm limit,'' and related expressions rise to $13.3\%$ of messages. Thus, SFT introduces useful bargaining moves and rhetorical forms, while PPO selects when to commit to them.

One undesirable behavior also appears here: the environment provides no external comparable-price information, so some learned market references are fabricated. We treat this learned bluffing behavior as a limitation rather than as evidence of improved factual grounding.

\paragraph{Deal-or-No-Deal: resist capitulation after rejection.}
In Deal-or-No-Deal, the main weakness of the base model appears after its initial proposal. Its openings are often reasonable, but rejection can trigger a sharp concession or repeated proposals that eventually lead to failure. For example, in one trajectory where hats and balls are each worth $5$ and books are worth $0$, the base model responds to resistance by accepting only a book, giving away all of its value. Other trajectories repeat nearly identical allocations until the counterpart walks away.

PPO makes rejection less likely to trigger this collapse. The trained policy more consistently preserves high-value items and either counters or accepts when the available allocation is favorable. Relative to the SFT diagnostic checkpoint, PPO also reduces the zero-reward and no-deal tail: zero-reward episodes fall from $4.7\%$ to $2.5\%$, and no-deals from $4.5\%$ to $0.5\%$. The qualitative shift is therefore concentrated in how the policy responds when its preferred allocation is challenged.

\paragraph{CaSiNo: maintain a floor on high-priority resources.}
CaSiNo exhibits a closely related pattern. Here, successful negotiation requires protecting high-priority resources while using low-priority resources as concessions. The base model often concedes too much after encountering resistance. In a matched example, it initially asks for all three food packages and, after one rejection, immediately drops to a single food package.

PPO learns a more stable reservation strategy. The trained policy explicitly maintains floors such as ``I require at least 2 food packages,'' preserves its high- and medium-priority resources, and gives ground primarily on its low-priority resource. At the distribution level, the fraction of over-conceding games falls from $17.3\%$ to $13.6\%$, while the fraction of games in which the agent strongly preserves its own value rises from $17.3\%$ to $24.2\%$. The SFT diagnostic moves in the opposite direction, raising over-concession to $29.4\%$, suggesting that imitation alone can favor agreeable behavior without calibrating how much value to surrender.

\paragraph{Job Interview: protect high-weight contract terms.}
Job Interview makes selective concession multi-dimensional. The two parties negotiate salary and holiday together with categorical terms such as position, workplace, and company. A strong policy therefore needs to identify which dimensions carry the largest private utility and spend concessions on less important issues.

The trained trajectories show greater preservation of high-value categorical terms. The worker's categorical utility in completed agreements rises from approximately $0.64$ for the base model to $0.73$--$0.78$ for the trained checkpoints, while the fraction of agreements securing the worker's top-choice workplace increases from $39\%$ to $68$--$72\%$. In matched examples, the base model can abandon several high-value categorical terms in order to close quickly, whereas the trained policy holds position, workplace, and company fixed while negotiating over salary and holiday.

\paragraph{Marketplace: create bargaining room and protect the reservation price.} Marketplace exhibits a particularly sharp strategic shift. The base buyer is already capable of completing transactions, but negotiates from an extremely weak position: its mean opening offer is $0.956$ of its private reservation price, leaving almost no room to bargain. After PPO, the mean opening falls to $0.361$ of the reservation price. The trained policy therefore begins far from its maximum willingness to pay and preserves substantial concession room.

Training also eliminates a severe information-leakage failure mode. In the base policy, $62\%$ of opening messages reveal the buyer's reservation price
or explicitly state its budget constraint. After PPO, this behavior essentially disappears. The learned shift closely parallels Craigslist: the agent stops revealing its private boundary, anchors aggressively, and moves toward agreement gradually rather than beginning near its own limit.

\paragraph{Calendar: optimize agreement quality rather than agreement rate.} Calendar exposes a different failure mode. The central challenge is not anchoring or bargaining over a scalar surplus, but selecting a mutually feasible time slot that also serves the principal's private preferences. Before Calendar training, the policy is highly willing to schedule a meeting but frequently accepts the counterpart's proposed slot without comparing it against better alternatives. Using the Marketplace-trained checkpoint as a pre-Calendar proxy, $92\%$ of games end in a scheduled meeting, yet only $16\%$ obtain a maximum-preference slot and $63\%$ receive zero utility.

Calendar PPO changes this objective. The trained policy schedules fewer meetings overall ($74\%$), but the quality of the selected slots improves substantially: the fraction of maximum-utility outcomes rises from $16\%$ to $38\%$, while zero-utility outcomes fall from $63\%$ to $35\%$. The policy has therefore learned that completing the coordination task is insufficient; it should reject or counter-propose when the available agreement poorly serves its principal.

\paragraph{Cross-domain pattern.}
Across all six environments, training makes agreement more explicitly conditional on the principal's utility. The base model often treats reaching an agreement as valuable in itself and gives ground too readily when the counterpart resists or proposes a feasible alternative. Post-training instead learns environment-specific reservation behavior: Craigslist and Marketplace create bargaining room through aggressive anchors and slower price concessions; DnD resists capitulation after rejection; CaSiNo protects high-priority resources; Job Interview preserves high-weight contract terms; and Calendar rejects feasible but low-value slots in favor of higher-preference alternatives.

The common strategic change is therefore \emph{utility-sensitive concession}: the policy becomes better at distinguishing what can be traded away from what should be protected. The particular implementation depends on the interaction structure: a reservation price in bilateral bargaining, a preferred bundle in allocation, high-weight issues in multi-issue negotiation, or a preference-weighted time slot in coordination.

\subsection{Interaction Behavior: Less Private Deliberation, More Strategic Communication}
\label{sec:qualitative:style}

The strategic changes above are accompanied by a broad shift in how the policy conducts the interaction. Table~\ref{tab:style} summarizes several trajectory-level measures. Across the four analyzed environments, trained policies generate substantially fewer total tokens per turn while sending longer and more informative messages to the counterpart. They also use rejection more actively and exhibit more explicit commitment language.

\begin{table}[t]
\centering
\small
\setlength{\tabcolsep}{4pt}
\begin{tabular}{@{}llccc@{}}
\toprule
\textbf{Env} & \textbf{Stage} & \textbf{Turns} & \textbf{Polite}
& \textbf{Firm} \\
\midrule
CaSiNo & untrained & 2.5 & 16\% & 1\% \\
       & SFT       & 3.4 & 10\% & 4\% \\
       & PPO       & 3.4 & \textbf{59\%} & \textbf{11\%} \\
\midrule
DnD    & untrained & 2.9 & 8\%  & 2\% \\
       & SFT       & 3.1  & 15\% & 23\% \\
       & PPO       & 3.2 & 36\% & 2\% \\
\midrule
Craigslist & untrained & 4.1 & 53\% & 3\% \\
           & SFT       & 5.4 & 15\% & 3\% \\
           & SFT$+$PPO & 4.6 & 51\% & \textbf{23\%} \\
\midrule
Job Interview & untrained & 4.6 & 35\% & 2\% \\
              & SFT       & 5.5 & 12\% & 7\% \\
              & SFT$+$PPO & 3.8 & 40\% & \textbf{43\%} \\
\bottomrule
\end{tabular}
\caption{\textbf{Interaction behavior across the training ladder.}
Turns denotes turns per game; public tok is the token length of the opponent-facing message; total tok includes the full generation associated with each turn; \texttt{<think>} is the fraction of turns containing an explicit reasoning block; polite and firm denote the fractions of public messages containing courteous and commitment language, respectively. Tokens are measured with the Qwen3-4B tokenizer.
}
\label{tab:style}
\end{table}

\paragraph{Rejection becomes a strategic action.}
The base policies rely heavily on proposing and accepting, with explicit rejection appearing in only $0$--$3\%$ of actions in the analyzed traces. Training makes rejection a meaningful part of the policy. Reject actions rise to approximately $14\%$ in DnD and $20\%$ in Job Interview, and they are typically accompanied by a counterproposal or explanation of which terms are unacceptable.

This change complements the selective-concession pattern above. The trained model has acquired a practical way to maintain a reservation boundary: instead of responding to an unfavorable proposal by immediately moving toward it, the policy can reject, explain the conflict, and propose an alternative.

\paragraph{Commitment becomes firmer while remaining socially cooperative.}
Training also changes the linguistic form of bargaining. Commitment language increases substantially in several environments: from $1\%$ to $11\%$ of messages in CaSiNo, $3\%$ to $23\%$ in Craigslist, and $2\%$ to $43\%$ in the analyzed Job Interview checkpoint. These messages contain explicit boundaries such as ``firm,'' ``final offer,'' or ``non-negotiable.'' Firmness generally coexists with courteous language. For example, politeness rises from $16\%$ to $59\%$ in CaSiNo and remains around half of Craigslist messages in the final policy. The learned behavior is therefore closer to \emph{polite commitment}: clear reservation boundaries expressed through cooperative language.

\paragraph{SFT broadens behavior; RL selects behavior by utility.}
The intermediate SFT checkpoints provide a useful diagnostic of how the two training signals differ. In Craigslist, SFT introduces capabilities that are almost absent from the base model: below-target anchoring, external justifications, and reduced leakage of the buyer's private target. RL then changes how these behaviors are deployed, producing stronger commitment and slower concession.

In the other environments where we evaluate SFT as an ablation, imitation can also introduce undesirable behavior: more looping in DnD, more over-concession in CaSiNo, and weaker categorical preservation in the currently analyzed Job Interview traces. Outcome optimization subsequently selects among these behaviors according to realized utility. This provides a trajectory-level interpretation of the training recipe: supervised data can expand the policy's behavioral repertoire, while reinforcement learning determines which parts of that repertoire are strategically useful.

\section{Conclusion}
\label{sec:conclusion}

We introduced \textsc{SocialRL}, a framework for training and studying language-model agents in heterogeneous delegated interactions. At the systems level, we develop a general event-based multi-agent environment interface that is independent of agent implementation, together with a decoupled training infrastructure that separates rollout generation from optimization through an OpenAI-compatible rollout proxy. These abstractions allow the same environments and agent harnesses to support local or remote agents, heterogeneous and black-box counterparts, reinforcement learning, and distillation without coupling the interaction stack to a particular trainer.

On top of this infrastructure, we study social-reasoning post-training across six heterogeneous negotiation and coordination environments. Domain-specific training produces strong 4B specialists whose aggregate performance lies in the range of much larger GPT models. Evaluating these specialists across all environments reveals substantial and highly directional transfer: structurally related interactions reinforce one another, while other training configurations introduce interference. We exploit this structure in two complementary approaches to unification. Transfer-aware cascade RL produces a single 4B policy with an Avg-6 of $0.627$, comparable to GPT-4.1, GPT-5.1, and GPT-5.2, and substantially outperforms random and anti-transfer curricula. MOPD provides a more efficient alternative, recovering most of the specialists' advantage in fewer than $100$ additional optimization steps.

We further investigate explicit theory-of-mind supervision as a complementary source of social-reasoning capability. Distilling \textsc{Infer}$\rightarrow$\textsc{Act}$\rightarrow$\textsc{Anticipate} traces improves both negotiation performance and measured theory-of-mind abilities, with next-action prediction emerging as the component most closely associated with negotiation outcomes. Trajectory analysis provides a consistent behavioral interpretation of these gains: trained agents become more sensitive to their principal's utility, protect private information more effectively, concede more selectively, and adapt their reasoning strategy to the structure of the interaction.

Taken together, the environment abstraction, decoupled training infrastructure, and training results provide a general platform for studying social reasoning in interactive agents. Our results suggest that strategic capabilities learned in one interaction structure can transfer to others, that this transfer can be exploited when consolidating specialists, and that strong multi-domain social behavior can be learned in relatively small models.

\bibliography{references} 
\bibliographystyle{iclr2024_conference}

\end{document}

%% file: math_commands.tex
\usepackage{amsmath,amsfonts,bm}

\def\eqref#1{equation~\ref{#1}}

\def\1{\bm{1}}

\DeclareMathAlphabet{\mathsfit}{\encodingdefault}{\sfdefault}{m}{sl}
\SetMathAlphabet{\mathsfit}{bold}{\encodingdefault}{\sfdefault}{bx}{n}

